\documentclass[dvipsnames,table]{article}

\usepackage[final]{hunyuan}

\usepackage{amsmath}
\usepackage{amssymb}
\usepackage{mathtools}
\usepackage{amsthm}
\usepackage{multirow}
\usepackage{makecell}
\usepackage{subcaption}
\usepackage{wrapfig}
\usepackage{graphicx}

\usepackage{pifont}
\usepackage{fvextra}
\usepackage{xspace}

\usepackage[utf8]{inputenc} % allow utf-8 input
\usepackage[T1]{fontenc}    % use 8-bit T1 fonts
\usepackage{url}            % simple URL typesetting
\usepackage{booktabs}       % professional-quality tables
\usepackage{amsfonts}       % blackboard math symbols
\usepackage{nicefrac}       % compact symbols for 1/2, etc.
\usepackage{microtype}      % microtypography
\usepackage{colortbl}
\usepackage{color, soul}
\usepackage{textgreek}

\usepackage[labelfont=bf]{caption}
\usepackage{enumitem}
\usepackage{sectsty}
\usepackage{pgfplots}
\usepackage{fancyhdr}
\usepackage{natbib} % 确保使用了natbib
\let\cite\citep    % 将 \cite 的定义指向 \citep

\renewcommand{\headrulewidth}{1pt}
\makeatletter
\def\headrule{{\if@fancyplain\let\headrulewidth\plainheadrulewidth\fi
\hrule\@height\headrulewidth\@width\textwidth \vskip-\headrulewidth}}
\makeatother

\definecolor{HYDarkBlue}{HTML}{2155EA} 
\definecolor{HYLightBlue}{HTML}{A8DFF6}
\definecolor{syh}{RGB}{41, 91, 160}
\usepackage[colorlinks, linkcolor=HYDarkBlue, anchorcolor=HYDarkBlue, citecolor=HYDarkBlue, urlcolor=HYDarkBlue]{hyperref}

\usepackage{cleveref}
\usepackage{threeparttable}

\sectionfont{\color{HYDarkBlue}\fontfamily{zi4}\selectfont}
\subsectionfont{\color{HYDarkBlue}\fontfamily{zi4}\selectfont}
\subsubsectionfont{\color{HYDarkBlue}\fontfamily{zi4}\selectfont}

\title{\method: Adaptive Verification Depth Pruning for Batched Speculative Decoding}

\author{%
\textbf{\normalsize{Tianyu Liu\textsuperscript{1},\quad Yuhao Shen\textsuperscript{2},\quad Rui Cen\textsuperscript{1},\quad Junhan Shi\textsuperscript{1},\quad Jiebin Zhang\textsuperscript{1}}}\\
\textbf{\normalsize{Guangshuo Qin\textsuperscript{1},\quad Hong Liu\textsuperscript{1},\quad Song Liu\textsuperscript{1},\quad Guanghua Yu\textsuperscript{1,*},\quad Jianchen Zhu\textsuperscript{1}}}\\[0.4em]
{\normalfont\small \textsuperscript{1}Tencent \quad \textsuperscript{2}Independent Researcher}\\
{\normalfont\small \textsuperscript{*}Corresponding author}
\vspace{0em}
}

\def\github{\raisebox{-1.5pt}{\includegraphics[height=1.05em]{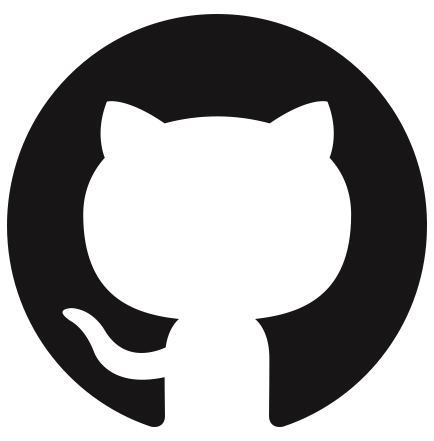}}}
\def\email{\raisebox{-1.5pt}{\includegraphics[height=1.05em]{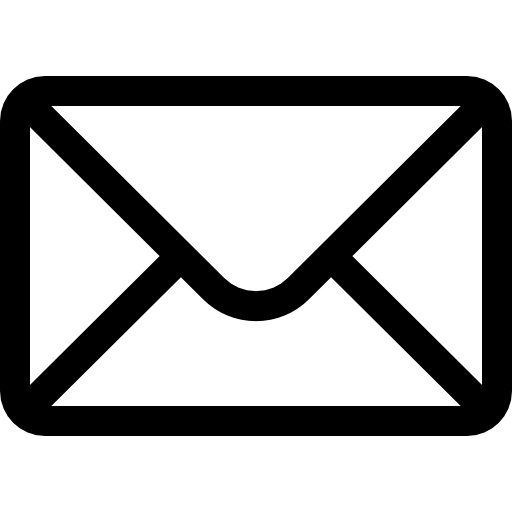}}}
\def\blog{\raisebox{-1.5pt}{\includegraphics[height=1.05em]{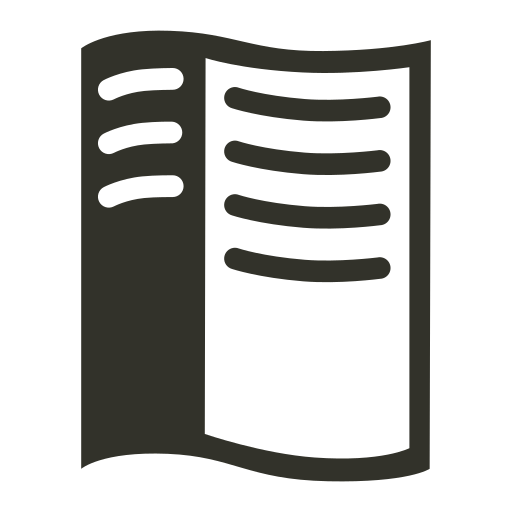}}}
\newcommand{\method}{\textbf{D-cut}\xspace}

\usepackage{geometry}
\usepackage{algorithm}
\usepackage{algpseudocode}
\usepackage{setspace}

\makeatletter
\renewcommand{\ALG@beginalgorithmic}{\small}
\makeatother



\begin{document}

\maketitle
\thispagestyle{fancy} % 恢复封面页的页眉和页脚

% \vspace{1em}
% v0
% \begin{abstract}
% Speculative decoding accelerates large language model (LLM) inference: a
% lightweight drafter proposes candidate tokens that the target model
% verifies in parallel. The recent state-of-the-art method DFlash replaces
% the autoregressive drafter with a lightweight block-diffusion model that
% generates an entire block of draft tokens in a single forward pass, making
% drafting nearly free and pushing single-request speedups beyond 6$\times$.
% However, we find that this advantage collapses under high concurrency:
% because DFlash proposes a large block of draft tokens at every step, the
% target model's verification cost grows rapidly with the batch size, and at
% a batch size of 16 DFlash is already slower than plain autoregressive
% decoding. A closer look reveals that only a small fraction of these draft
% tokens are ultimately accepted, while the majority are verified in vain.
% Motivated by this, we propose \method, which dynamically prunes
% DFlash's draft tokens so that the limited verification budget is spent on
% the tokens most likely to be accepted. \method prunes adaptively along
% two axes: (1)~\emph{cross-request pruning}, which allocates the
% verification budget across concurrent requests, and (2)~\emph{runtime
% adaptation}, which adjusts the pruning ratio to the underlying hardware and
% batch size. Experiments on \comments{[benchmark]} show that \method
% achieves up to \comments{[X]$\times$} speedup over DFlash under high
% concurrency, while preserving the target model's output distribution.
% \end{abstract}

% v1
\begin{abstract}
    Speculative decoding has become a leading approach for accelerating large language model (LLM) inference without compromising output quality.
    Recent methods further improve single-request speedup by decoupling draft length from drafting latency via parallel generation, which produces longer drafts to achieve higher mean accepted tokens (MAT).
    However, as request concurrency grows, these long drafts waste more compute on rejected tokens, inflating the \emph{verification cost} and making speculative decoding suboptimal in high-concurrency scenarios.
    To address this challenge, we propose \method, an adaptive pruning strategy that selects draft tokens across the entire batch, concentrating verification budget on tokens most likely to be accepted.
    Our \method is motivated by two observations:
    (1) acceptance length varies significantly across concurrent requests, so \method performs \emph{cross-request pruning} that adaptively allocates verification budget based on draft confidence;
    (2) verification cost largely depends on the runtime environment (GPU, parallelism, etc.), so we incorporate a \emph{runtime cost model} into the selection, allowing \method to adapt its pruning depth to the actual deployment.
    % Experiments across dense and MoE models show that, at high concurrency where the long-draft baseline can degrade below $1\times$, \method lifts the average speedup from $1.26\times$ to $1.65\times$: it restores dense models that had fallen below the autoregressive baseline, and on MoE models it reaches up to $3.0\times$ over autoregressive decoding.
    Experiments across dense and MoE models show that \method improves the average speedup from $1.26\times$ to $1.65\times$ under high concurrency, restores acceleration for dense-model configurations where long-draft baselines fall below autoregressive decoding, and achieves up to $3.0\times$ speedup over autoregressive decoding on MoE models.
\end{abstract}

\section{Introduction}
\label{sec:introduction}

Large language models (LLMs) have demonstrated strong capabilities across a wide range of tasks~\citep{gpt-5-5, deepseek-v4, qwen35blog}.
However, their autoregressive decoding (AR) paradigm, which generates only one token per forward pass, fundamentally limits inference throughput, as each step requires loading the full model parameters from memory~\citep{palm-scaling}.
This memory-bandwidth bottleneck has motivated extensive research into accelerating LLM inference.

Speculative decoding~\citep{sps1, sps2} has emerged as a leading lossless acceleration approach.
The core idea is \emph{draft-then-verify}: a lightweight drafter proposes multiple candidate tokens, which the target model verifies in a single forward pass.
In low-concurrency settings, verification can process many tokens in parallel at roughly the same cost as generating one, so each accepted draft token effectively saves a full decoding step.
The speedup is governed by two factors: the \emph{acceptance length} (how many draft tokens are accepted per step) and the \emph{step time} (how expensive verification is per step)~\citep{xia2024survey}.

Recent research has sought to improve speculative decoding by increasing the draft length and acceptance rate while minimizing drafting latency.
Early methods rely on independent small models as drafters~\citep{sps1, sps2, pearl, Zhang_2024}.
To improve draft accuracy without introducing excessive overhead, subsequent work proposes reusing the target model's hidden features to drive lightweight, single-layer autoregressive drafters~\citep{eagle, glide, medusa, mtp}.
More recently, this feature-reuse approach has evolved into block-parallel drafters that generate an entire block of draft tokens in a single forward pass, effectively decoupling draft length from drafting latency~\citep{dflash, p-eagle, huang2026domino,zhang2026dflare}.
With parallel drafting, these block-wise drafters generate longer drafts and achieve higher mean accepted tokens (MAT), leading to better speedups at low concurrency.
% These advances have pushed single-request speedups well beyond $3\times$, with state-of-the-art methods like DFlash~\citep{dflash} achieving over $6\times$ through block-parallel diffusion drafting.

\begin{figure}[t]
    \centering
    \begin{subfigure}[t]{0.45\textwidth}
        \centering
        \includegraphics[width=\linewidth]{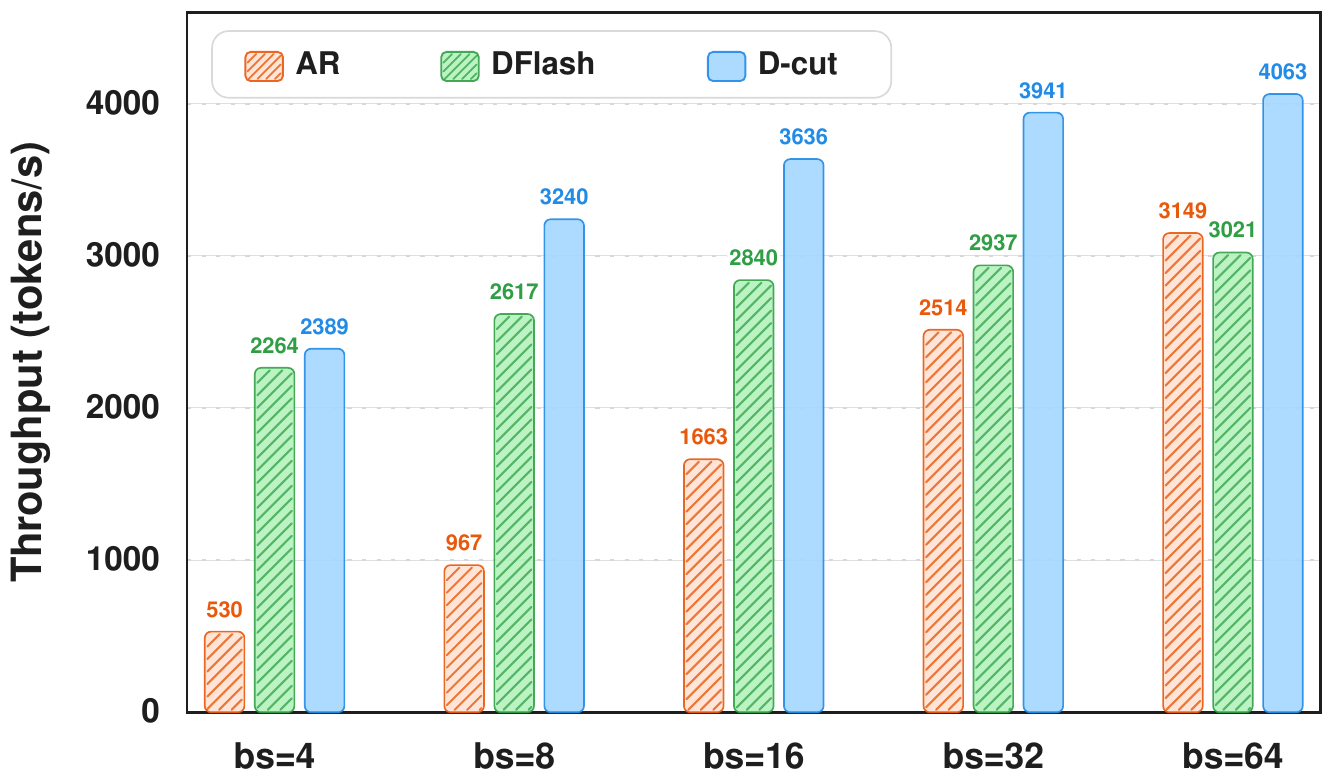}
        \caption{Throughput under increasing batch size.}
        \label{fig:intro-throughput}
    \end{subfigure}
    \hfill
    \begin{subfigure}[t]{0.53\textwidth}
        \centering
        \includegraphics[width=\linewidth]{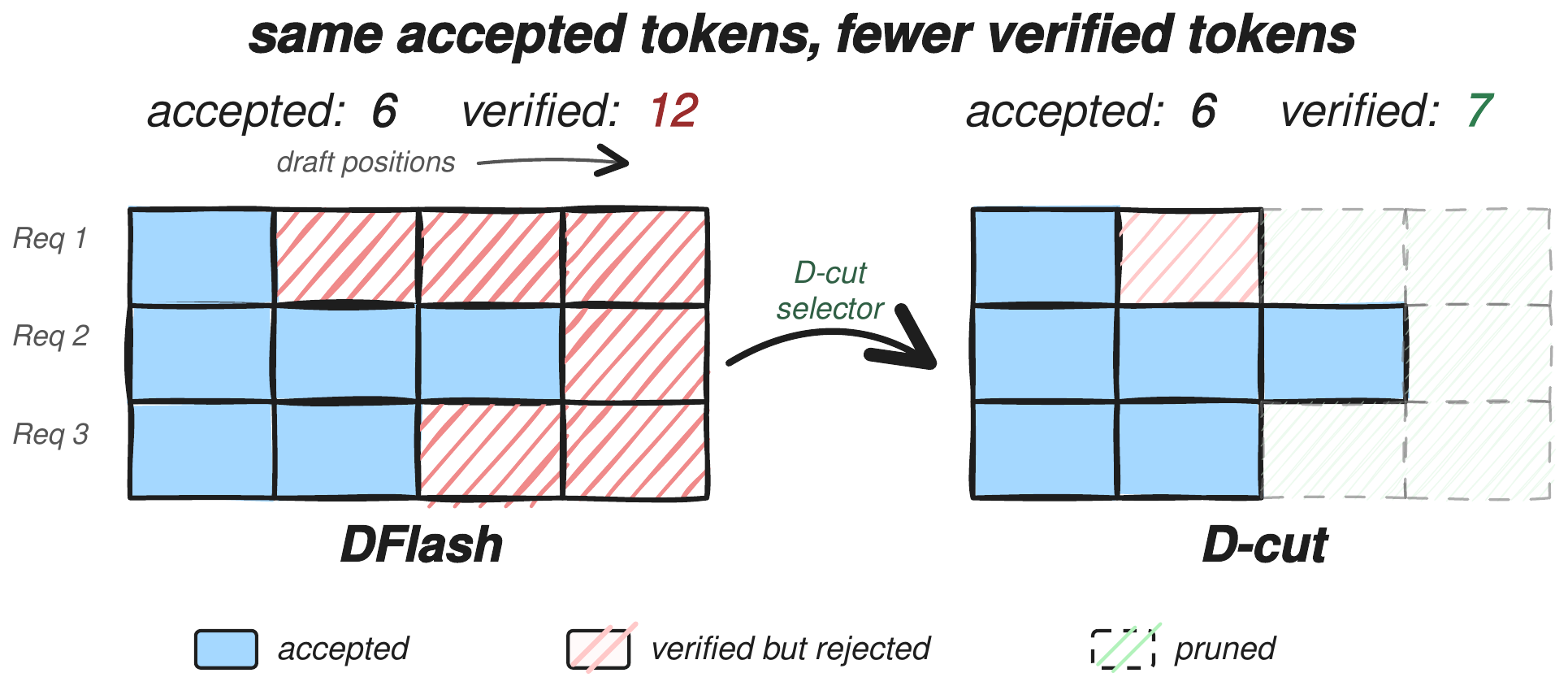}
        \caption{Cross-request verification-depth pruning.}
        \label{fig:intro-pruning}
    \end{subfigure}
    \caption{\textbf{Motivation and overview of \method.} (a) Throughput on Math500 with Qwen3-8B on a single H20 GPU. DFlash benefits from long block-parallel drafts at small batch sizes, but its throughput saturates as batched verification becomes compute-bound and eventually falls below standard AR decoding at high concurrency (bs=64). (b) \method treats all draft tokens in a batch as sharing one verification budget. Instead of verifying the same draft depth for every request, it ranks drafts by confidence and prunes low-confidence suffixes, reallocating verification compute toward requests whose drafts are more likely to be accepted.}
    \label{fig:intro}
\end{figure}

However, \textbf{as request concurrency grows, the speedup from speculative decoding shrinks and can even fall below standard autoregressive decoding}~\citep{illusion}.
In batched serving, the target model must verify $B \times (D+1)$ tokens per step, where $B$ is the batch size and $D$ is the draft length.
As the batch size increases, verification transitions from memory-bound to compute-bound, and the cost of processing each additional draft token becomes non-negligible~\citep{echo}.
This cost growth is compounded by low draft utilization: for DFlash with $D=15$, the average acceptance length is only $\sim$6 tokens, meaning roughly 60\% of verified draft tokens are rejected.
\Cref{fig:intro-throughput} quantifies this effect: DFlash delivers large speedups at small batch sizes, but as concurrency increases, its throughput saturates while AR keeps scaling, until at $\text{bs}=64$ it falls below standard autoregressive decoding.
% In contrast, \method selectively prunes low-utility draft tokens before verification, sustaining higher throughput across the concurrency range.

To address this challenge, we propose \method, an adaptive pruning strategy that selectively retains draft tokens across the entire batch, concentrating verification compute on positions most likely to be accepted.
As illustrated in \Cref{fig:intro-pruning}, \method treats all draft tokens in a batch as sharing one verification budget and prunes low-confidence drafts per request.
It is built on two mechanisms:
First, \textbf{cross-request pruning} relaxes the uniform-depth constraint. Given the significant variation in acceptance lengths across a batch, \method leverages the drafter's prediction confidence to globally rank all candidate tokens, shifting verification budget from low-confidence suffixes to high-confidence continuations across requests.
Second, \textbf{runtime-adaptive depth selection} aligns the pruning strategy with actual runtime conditions. Recognizing that the latency penalty of verifying extra tokens depends heavily on the runtime environment, including model architecture, parallelism strategy, and GPU type, \method profiles this cost at startup. It then incorporates the runtime-specific cost into its selection objective to choose a suitable pruning depth for the current deployment.

Our main contributions are as follows:
\begin{itemize}[leftmargin=1.5em, itemsep=2pt]
    % \item We identify the \emph{verification cost explosion} problem in batched speculative decoding, showing that DFlash degrades below the autoregressive baseline at high concurrency.
    \item We identify the \emph{verification cost explosion} problem in batched speculative decoding, showing that DFlash can fall below the autoregressive baseline at high concurrency. This exposes a mismatch between cheap long-draft generation and costly batched verification of rejected tokens.
    \item We propose \method, which performs cross-request pruning guided by draft confidence and runtime-adaptive depth selection via a profiled cost model. \method is training-free, requires no modification to the target model, and preserves the lossless output distribution.
    \item We evaluate \method on both dense and MoE models across a range of batch sizes, demonstrating that it recovers acceleration for dense configurations where vanilla DFlash falls below the autoregressive baseline at high concurrency, and reaches up to $3.0\times$ speedup over autoregressive decoding on MoE models.
\end{itemize}

\section{Motivation}
\label{sec:motivation}

\begin{figure}[t]
    \centering
    \begin{subfigure}[t]{0.49\textwidth}
        \centering
        \includegraphics[width=\linewidth]{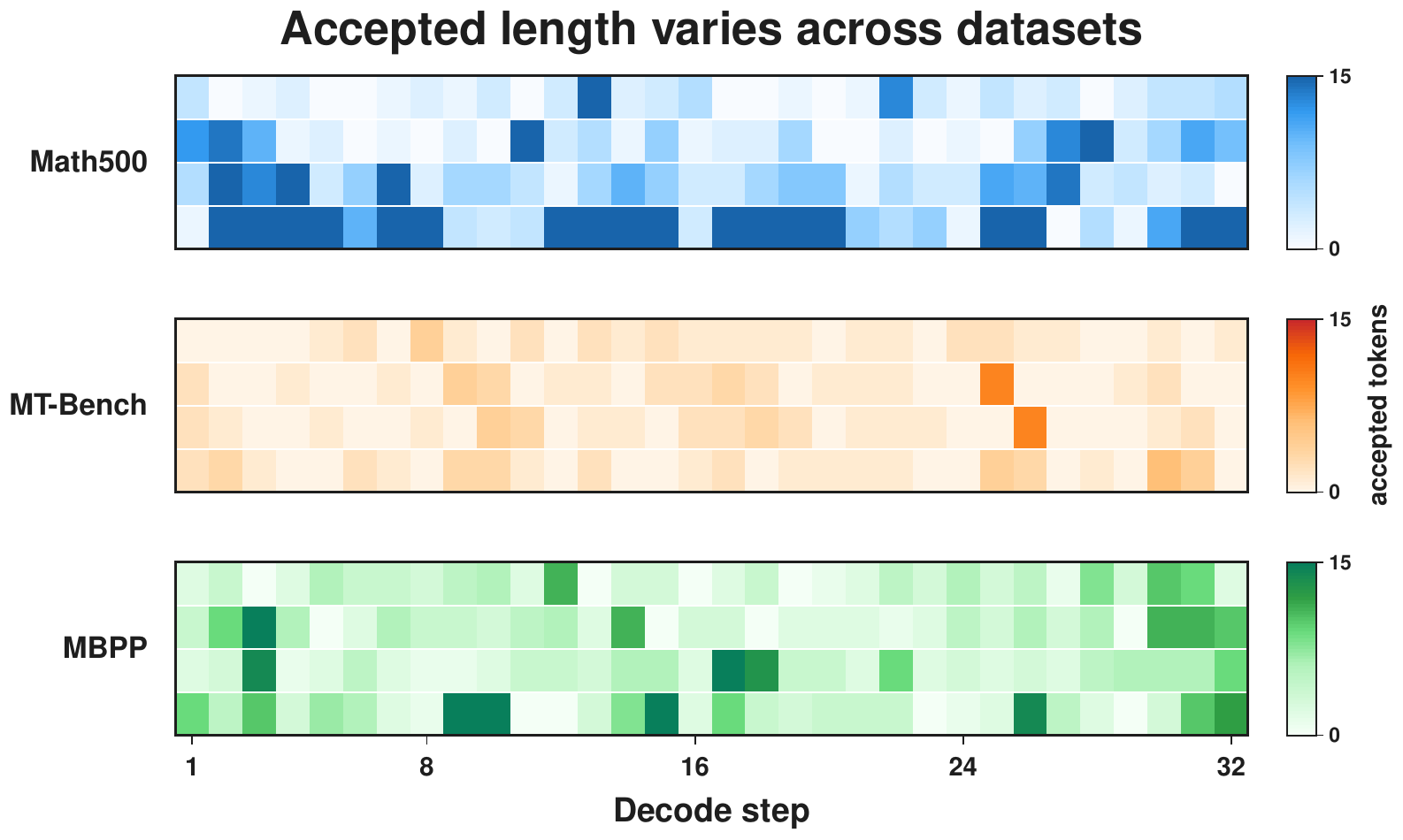}
        \caption{Accepted-length variation across datasets.}
        \label{fig:motivation-mat-variation}
    \end{subfigure}
    \hfill
    \begin{subfigure}[t]{0.49\textwidth}
        \centering
        \includegraphics[width=\linewidth]{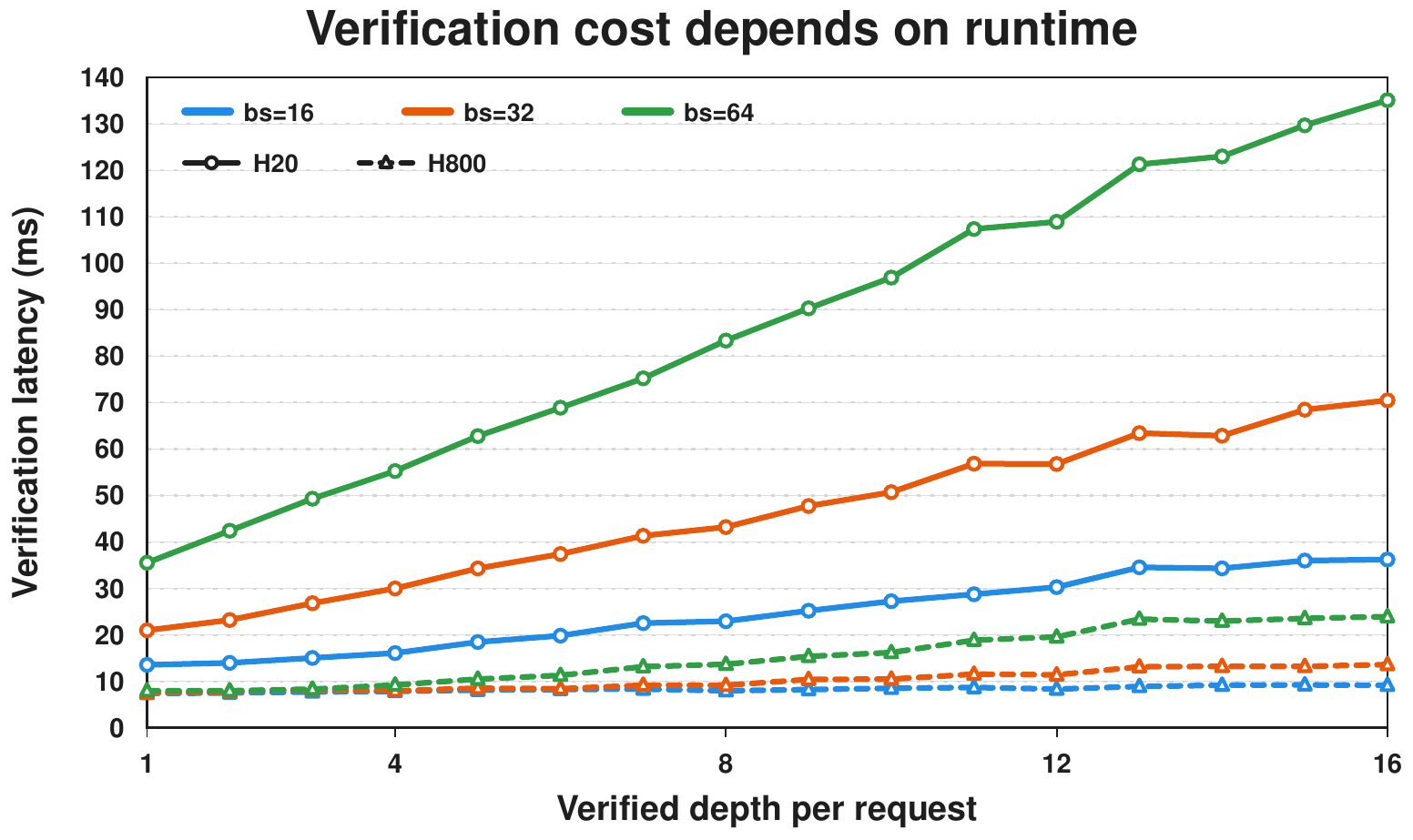}
        \caption{Runtime verification cost profiling.}
        \label{fig:motivation-runtime-cost}
    \end{subfigure}
    \caption{\textbf{Motivating observations for \method.} (a) MAT trajectories for four representative requests in DFlash, using Qwen3-8B and draft length $D{=}15$. Accepted length varies substantially across requests, decoding steps, and datasets, motivating \textit{cross-request allocation} of the verification budget. (b) P50 verification latency when varying the verified depth per request on H20 and H800. The cost curves change with batch size, GPU type, model architecture, and parallelism strategy, motivating \textit{runtime-adaptive depth selection}.}
    \label{fig:motivation}
\end{figure}

\subsection{Accepted Length Varies Across Requests}
\label{sec:motivation-cross-request}

Recent studies on autoregressive drafters have established that the number of accepted draft tokens varies significantly across requests and decoding steps~\citep{pearl,talon,double}.
A natural question arises: \textit{does this high variance persist for block-parallel diffusion drafters like DFlash?}
To investigate this, we tracked the accepted length of four representative requests during a stable high-concurrency window, using DFlash with draft length $D{=}15$ on Qwen3-8B.

\Cref{fig:motivation-mat-variation} shows that block-parallel drafters exhibit the same high variance.
Within a single decoding step, one request may accept more than 12 draft tokens while another accepts fewer than 3, yet both pay the same verification cost of $D{+}1{=}16$ positions.
The same request swings between long and short acceptance across steps, and the overall pattern varies by benchmark: Math500~\citep{math500} and MBPP~\citep{mbpp} produce frequent long-acceptance sequences, while MT-Bench~\citep{mt-bench} is dominated by short prefixes where most full-depth verification is wasted.
Moreover, similar to autoregressive drafters~\citep{eagle2, opt-tree}, the block-diffusion drafter's own prediction confidence provides a lightweight proxy for whether a draft token will be accepted by the target model. (Validated in \Cref{sec:appendix-confidence}.)

Despite this variance, block-parallel methods such as DFlash and P-EAGLE~\citep{p-eagle} verify a fixed draft depth for every request in the batch.
Such a fixed depth must be set conservatively to accommodate long-acceptance requests, wasting target-model compute on the low-yield suffixes of others; verification is better allocated at the granularity of individual draft tokens.
Rather than a uniform depth, \method uses the drafter's confidence to globally rank candidate tokens across the batch, retaining high-confidence drafts while pruning low-confidence suffixes.

\subsection{Verification Cost Depends on Runtime}
\label{sec:motivation-runtime-cost}

Having established \emph{which} tokens to verify, the second question is \emph{how much} to prune.
A natural intuition is that the right amount of pruning is tied to the runtime environment, because the cost of verifying extra draft tokens differs sharply across deployments.
\Cref{fig:motivation-runtime-cost} profiles P50 verification latency while varying the verified depth per request from 1 to 16.
On the H20, a compute-limited GPU, latency rises steeply with depth: at batch size 16, raising the depth from 4 to 16 adds about 20\,ms per step (from $16$ to $36$\,ms), so pruning should be aggressive to avoid inflating step time.
On a compute-rich GPU such as the H800, the same range adds barely 1\,ms (from $8$ to $9$\,ms) and the curve stays nearly flat, so verifying extra tokens is cheap and pruning should be conservative.
Model architecture (dense vs.\ MoE), parallelism strategy (TP/DP), and batch size shift these cost curves in the same way~\citep{sequoia, magicdec}.
A single fixed pruning strategy can therefore work well in one environment yet perform poorly in another.

A better solution is to model this runtime cost explicitly and prune against it.
The right quantity to maximize is the speedup over autoregressive (AR) decoding, which factors into a benefit term and a cost term.
Writing the verification budget as $K$, the number of block positions retained for verification across the batch, the speedup at a given batch size is
\begin{equation}
    \mathrm{Speedup}(K)
    = \frac{\mathrm{MAT}(K)}{T_{\mathrm{spec}}(K) / T_{\mathrm{AR}}},
    \label{eq:motivation-speedup}
\end{equation}
where $\mathrm{MAT}(K)$ is MAT per step, $T_{\mathrm{spec}}(K)$ is the full latency of one speculative step (drafting, selection, verification and others), and $T_{\mathrm{AR}}$ is the latency of one AR decoding step.
\textbf{This formulation naturally decouples the speedup into two orthogonal factors:}
The numerator $\mathrm{MAT}(K)$ captures the algorithmic benefit, which depends on the drafter's accuracy and is independent of the hardware; it can be estimated dynamically using confidence scores (\Cref{fig:confidence-acceptance}).
The overhead $T_{\mathrm{spec}}(K)$ in the denominator is purely a hardware characteristic, growing with the verification budget $K$ but independent of the drafter's acceptance rate.
Because the algorithmic benefit and hardware overhead are decoupled, they can be modeled separately.
\method estimates the benefit via draft confidence at runtime and profiles the hardware cost curve once at startup, combining them to dynamically find the optimal budget $K$ that maximizes \Cref{eq:motivation-speedup}.

\section{Method}
\label{sec:method}

\method reduces the verification cost of block-parallel speculative decoding by framing draft pruning as a dynamic budget allocation problem.
As identified in \Cref{sec:motivation}, an optimal pruning strategy must account for both the highly variable acceptance lengths across requests and the specific runtime verification costs.
We describe the method based on DFlash, though the principles apply to any batched block-parallel drafter.

\begin{figure}[t]
    \centering
    \includegraphics[width=\linewidth]{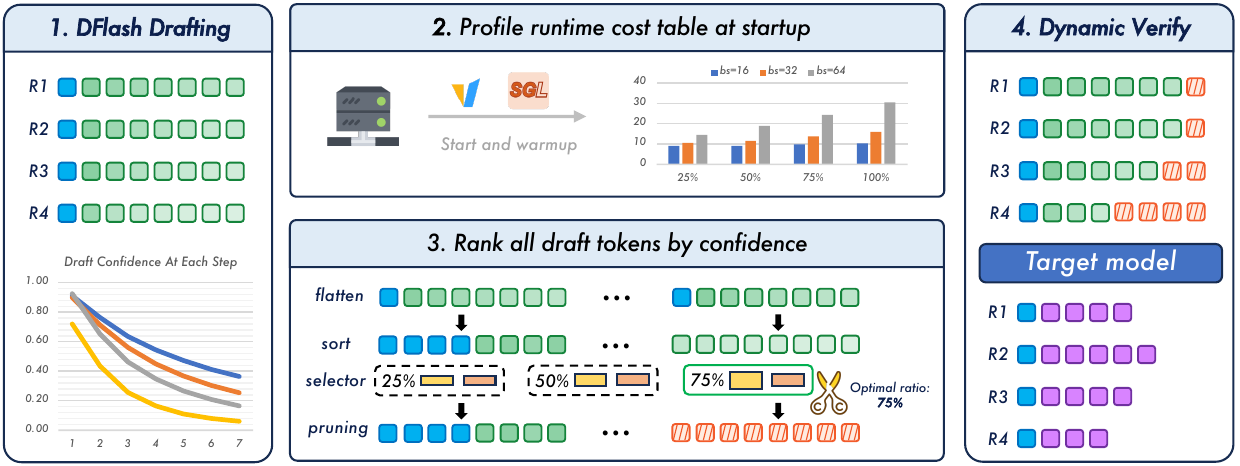}
    \caption{\textbf{Overview of \method.} The drafter produces batched draft blocks with token-level confidence scores. \method evaluates the expected utility of each position, dynamically selects a runtime-aware global budget, and packs only the top candidates into a dense verification batch for the target model.}
    \label{fig:method-overview}
\end{figure}

\subsection{Problem Setup and Pipeline Overview}
\label{sec:method-setup}

Consider a speculative decoding step with a batch of $B$ active requests.
For each request $i$, the drafter generates a block of $D$ draft tokens $z_{i,1:D}$ in a single forward pass ($D=15$ in our implementation).
These drafts are appended to the \emph{bonus token} $z_{i,0}$ from the previous step (the token just accepted by the target model, on which the drafter conditions), forming a verification candidate block of $D{+}1$ positions.
Because the bonus token is guaranteed to be accepted, verifying it ensures at least one token of progress per request.
Rather than verifying all $D$ draft tokens for every request, \method adaptively truncates the block, submitting a prefix of length $n_i{+}1 \le D{+}1$ (the bonus token plus $n_i$ drafts) to the target model.

As illustrated in \Cref{fig:method-overview}, \method determines these keep depths $n_i$ through a two-stage process.
First, it scores every candidate position in the batch by estimating its expected utility, allowing high-confidence drafts to be ranked above low-confidence ones (\Cref{sec:method-utility}).
Second, it evaluates the aggregated utility of the batch against a pre-profiled hardware cost curve, dynamically selecting a global verification budget $K$ that maximizes the projected step speedup (\Cref{sec:method-runtime}).
The top-$K$ scoring candidates are then packed and sent to the target model, preserving the exact target distribution while minimizing wasted compute.

\subsection{Estimating Verification Utility}
\label{sec:method-utility}

To globally rank candidates across the batch, \method must quantify the expected benefit of verifying each position.
Let $L_i \in \{0,\ldots,D\}$ be the number of draft tokens the target model would accept if the full block were verified.
Because acceptance is strictly sequential, keeping the first $n_i$ draft tokens advances a total of $A_i(n_i) = 1 + \min(L_i, n_i)$ tokens (the certain bonus plus the accepted draft prefix).
The expected number of tokens advanced under keep depth $n_i$ is entirely determined by the block's survival probabilities:
\begin{equation}
    \mathbb{E}[A_i(n_i)] = 1 + \sum_{k=1}^{n_i} \Pr(L_i \ge k).
    \label{eq:dcut-expected-accepted}
\end{equation}

\method estimates these survival probabilities using the drafter's own prediction confidence.
Let $c_{i,t} \in [0,1]$ be the drafter's confidence for the token at draft position $t$.
Since position $k$ is reached only if all its preceding tokens are accepted, the probability that the block survives through depth $k$ is given by the prefix product:
\begin{equation}
    s_{i,k} = \prod_{t=1}^{k} c_{i,t}, \quad \text{with } s_{i,0}=1 \text{ for the certain bonus token.}
    \label{eq:dcut-prefix-score}
\end{equation}
Substituting $s_{i,k}$ into \Cref{eq:dcut-expected-accepted} yields an estimated advance of $\hat{A}_i(n_i) = \sum_{k=0}^{n_i} s_{i,k}$.
Each term $s_{i,k}$ elegantly isolates the expected marginal token advance from verifying position $k$.
Because $s_{i,k}$ is monotonically decreasing with depth $k$, any global top-$K$ selection over these scores will naturally yield a contiguous prefix for each request.

\subsection{Runtime-Adaptive Budget Allocation}
\label{sec:method-runtime}

With every position scored, \method must decide the global verification budget $K$: the total number of positions to verify across the batch.
A fixed $K$ is suboptimal across different hardware (\Cref{sec:motivation-runtime-cost}), so \method dynamically selects $K$ to maximize the overall speedup.

To avoid the memory explosion of capturing a separate CUDA graph for every possible arbitrary $K$, \method restricts the budget to a small set of discrete ratios $\mathcal{R}=\{0.25, 0.50, 0.75, 1.00\}$.
For a given ratio $\rho \in \mathcal{R}$ and batch size $B$, the allowed capacity is
\begin{equation}
    K_{\rho}(B) = \max\!\left(B,\ \left\lceil \rho \cdot B(D+1) \right\rceil \right),
    \label{eq:dcut-budget}
\end{equation}
where the lower bound $B$ ensures every request retains at least its bonus token.
At server startup, \method profiles the hardware to build a latency cost table $C(B,\rho)$, which measures the end-to-end time of a speculative step for each discrete batch shape.

During serving, \method flattens all $B(D{+}1)$ candidate positions and sorts them by $s_{i,k}$.
For each candidate ratio $\rho$, it sums the scores of the top-$K_{\rho}(B)$ positions, representing the expected tokens advanced by the entire batch under that budget.
This sum directly estimates the algorithmic benefit (the numerator) from the speedup objective in \Cref{eq:motivation-speedup}.
For the cost denominator, the speculative step latency $T_{\mathrm{spec}}$ is provided by the profiled table $C(B,\rho)$.
While the strict speedup denominator is $T_{\mathrm{spec}} / T_{\mathrm{AR}}$, the autoregressive latency $T_{\mathrm{AR}}$ is constant for a given batch size $B$ and independent of the chosen ratio $\rho$.
Because it does not affect the argmax operation, $T_{\mathrm{AR}}$ is omitted.
\method therefore selects the optimal ratio $\rho^\star$ by maximizing:
\begin{equation}
    \rho^\star = \arg\max_{\rho \in \mathcal{R}} \frac{\sum_{(i,k) \in \mathcal{S}_{K_{\rho}(B)}} s_{i,k}}{C(B,\rho)},
    \label{eq:dcut-objective}
\end{equation}
where $\mathcal{S}_{K_{\rho}(B)}$ is the selected top-$K_{\rho}(B)$ set.
By dynamically choosing $\rho^\star$, \method ensures that high-confidence drafts are aggressively verified when compute is cheap (e.g., on an H800), but heavily pruned when verification is expensive (e.g., on an H20) or when draft quality is poor.
Crucially, because \method only drops low-utility suffixes and leaves the target model's accept/reject logic unaltered, the verified output exactly matches the target model's distribution (\Cref{sec:appendix-execution}).

\section{Experiments}
\label{sec:experiments}

\subsection{Experimental Setup}
\label{sec:experiments-setup}

\paragraph{Models and benchmarks.}
We evaluate six target models that cover both dense and MoE architectures: Llama-3.1-8B~\citep{llama3}, Qwen3-4B, Qwen3-8B, Qwen3.5-27B, Qwen3.5-35B-A3B~\citep{qwen35blog}, and Hy3-295B-A21B~\citep{hy3}.
Our evaluation benchmarks span math reasoning, code generation, and multi-turn chat: GSM8K~\citep{gsm8k}, Math500~\citep{math500}, HumanEval~\citep{humaneval}, MBPP~\citep{mbpp}, and MT-Bench~\citep{mt-bench}.
Unless otherwise stated, we use greedy target decoding with greedy draft proposals and target-only verification, sweep the serving concurrency over $\{4, 8, 16, 32, 64\}$, and report main results at concurrency 32; \Cref{sec:experiments-temperature} changes only the target sampling temperature to 1.

\paragraph{Baselines and implementation.}
Our primary baseline is DFlash~\citep{dflash}, which uses a block-diffusion drafter to propose an entire block of draft tokens in one forward pass. We evaluate two block sizes 8 and 16, denoted as $(\textbf{8})$ and $(\textbf{16})$ respectively.
For broader context across SOTA methods, we further benchmark EAGLE-3~\citep{eagle3} on the 4B/8B targets, MTP~\citep{qwen35blog,hy3} on the Qwen3.5 and Hy3 targets using each model's default configuration, and standard autoregressive (AR) decoding as the throughput reference.
All methods run on a single node of 8$\times$ NVIDIA H20 GPUs; the cross-hardware study in \Cref{sec:experiments-ablation} repeats the Qwen3-8B block-16 sweep on 8$\times$ NVIDIA H800 GPUs. Serving and implementation details are deferred to \Cref{sec:appendix-details}.
\paragraph{Metrics.}
We use two metrics in our experiments: Mean accepted tokens (MAT, denoted $\tau$) and throughput speedup (denoted $Spd.$) over the AR baseline.
We treat throughput speedup as the primary metric: under batched serving a longer draft raises MAT but also enlarges the verification batch, so a high MAT does not necessarily translate into higher throughput~\citep{talon}. Reporting both separates the algorithmic benefit of pruning from the serving efficiency it actually realizes.

\subsection{Main Results}
\label{sec:experiments-main}

\begin{table*}[t]
    \centering
    \small
    \setlength{\tabcolsep}{5pt}
    \renewcommand{\arraystretch}{1.25}
    \setlength{\aboverulesep}{0pt}
    \setlength{\belowrulesep}{0pt}
    \def\hl{\cellcolor{HYLightBlue!55}}
    \definecolor{HYLightGreen}{HTML}{CDEBD6}
    \def\hlg{\cellcolor{HYLightGreen}}
    \caption{\textbf{End-to-end serving results at concurrency 32.} $\tau$: mean accepted tokens (MAT). Spd.: output-token throughput speedup over autoregressive decoding. Avg.: averages over the five datasets. \textbf{Bold} marks the best speedup per column within each model. Shaded rows denote \method, green for the (8) budget and blue for the (16) budget.}
    \label{tab:main-results}
    \resizebox{\textwidth}{!}{%
    \begin{tabular}{cl*{6}{cc}}
        \toprule
        \multirow{2}{*}{\textbf{Model}} & \multirow{2}{*}{\textbf{Method}}
            & \multicolumn{2}{c}{\textbf{GSM8K}}
            & \multicolumn{2}{c}{\textbf{Math500}}
            & \multicolumn{2}{c}{\textbf{HumanEval}}
            & \multicolumn{2}{c}{\textbf{MBPP}}
            & \multicolumn{2}{c}{\textbf{MT-Bench}}
            & \multicolumn{2}{c}{\textbf{Avg.}} \\
        \cmidrule(lr){3-4}\cmidrule(lr){5-6}\cmidrule(lr){7-8}\cmidrule(lr){9-10}\cmidrule(lr){11-12}\cmidrule(lr){13-14}
        & & $\tau$ & Spd. & $\tau$ & Spd. & $\tau$ & Spd. & $\tau$ & Spd. & $\tau$ & Spd. & $\tau$ & Spd. \\
        \midrule
        \multirow{5}{*}{\shortstack{Llama-3.1\\8B}}
          & EAGLE-3 & 2.41 & 0.87$\times$ & 1.26 & 0.66$\times$ & 2.47 & 1.03$\times$ & 3.14 & \textbf{1.62$\times$} & 2.23 & 0.97$\times$ & 2.30 & 1.03$\times$ \\
          & DFlash (\textbf{8})  & 4.32 & 0.91$\times$ & 4.72 & 1.25$\times$ & 4.58 & 1.00$\times$ & 4.75 & 1.30$\times$ & 3.83 & 1.04$\times$ & 4.44 & 1.10$\times$ \\
          & \hlg\method (\textbf{8})  & \hlg 4.08 & \hlg\textbf{1.13$\times$} & \hlg 4.41 & \hlg\textbf{1.48$\times$} & \hlg 4.30 & \hlg\textbf{1.12$\times$} & \hlg 4.39 & \hlg 1.54$\times$ & \hlg 3.49 & \hlg\textbf{1.34$\times$} & \hlg 4.13 & \hlg\textbf{1.32$\times$} \\
          & DFlash (\textbf{16}) & 4.46 & 0.56$\times$ & 4.96 & 0.74$\times$ & 4.80 & 0.66$\times$ & 5.06 & 0.77$\times$ & 3.94 & 0.60$\times$ & 4.64 & 0.67$\times$ \\
          & \hl\method (\textbf{16}) & \hl 3.93 & \hl 0.94$\times$ & \hl 4.69 & \hl 1.25$\times$ & \hl 4.33 & \hl 0.98$\times$ & \hl 4.33 & \hl 1.30$\times$ & \hl 3.48 & \hl 1.13$\times$ & \hl 4.15 & \hl 1.12$\times$ \\
        \midrule
        \multirow{5}{*}{\shortstack{Qwen3\\4B}}
          & EAGLE-3 & 2.48 & 1.07$\times$ & 2.36 & 1.44$\times$ & 2.34 & 1.28$\times$ & 2.26 & 1.42$\times$ & 2.11 & \textbf{1.21$\times$} & 2.31 & 1.28$\times$ \\
          & DFlash (\textbf{8})  & 4.85 & 1.16$\times$ & 5.67 & 1.67$\times$ & 4.88 & 1.31$\times$ & 4.53 & 1.46$\times$ & 2.84 & 0.84$\times$ & 4.55 & 1.29$\times$ \\
          & \hlg\method (\textbf{8})  & \hlg 4.67 & \hlg\textbf{1.33$\times$} & \hlg 5.34 & \hlg\textbf{1.77$\times$} & \hlg 4.71 & \hlg\textbf{1.44$\times$} & \hlg 4.38 & \hlg\textbf{1.58$\times$} & \hlg 2.73 & \hlg 1.06$\times$ & \hlg 4.37 & \hlg\textbf{1.44$\times$} \\
          & DFlash (\textbf{16}) & 6.14 & 0.90$\times$ & 7.87 & 1.38$\times$ & 6.41 & 0.99$\times$ & 5.88 & 1.08$\times$ & 3.23 & 0.56$\times$ & 5.91 & 0.98$\times$ \\
          & \hl\method (\textbf{16}) & \hl 5.80 & \hl 1.22$\times$ & \hl 6.95 & \hl 1.72$\times$ & \hl 5.98 & \hl 1.35$\times$ & \hl 5.48 & \hl 1.47$\times$ & \hl 3.05 & \hl 0.99$\times$ & \hl 5.45 & \hl 1.35$\times$ \\
        \midrule
        \multirow{5}{*}{\shortstack{Qwen3\\8B}}
          & EAGLE-3 & 2.31 & 0.96$\times$ & 2.26 & 1.19$\times$ & 2.26 & 1.08$\times$ & 2.08 & 1.25$\times$ & 1.95 & 1.08$\times$ & 2.17 & 1.11$\times$ \\
          & DFlash (\textbf{8})  & 4.87 & 0.99$\times$ & 5.69 & 1.42$\times$ & 5.06 & 1.14$\times$ & 4.61 & 1.26$\times$ & 2.91 & 0.78$\times$ & 4.63 & 1.12$\times$ \\
          & \hlg\method (\textbf{8})  & \hlg 4.47 & \hlg\textbf{1.21$\times$} & \hlg 5.20 & \hlg\textbf{1.51$\times$} & \hlg 4.51 & \hlg\textbf{1.26$\times$} & \hlg 4.07 & \hlg\textbf{1.50$\times$} & \hlg 2.81 & \hlg\textbf{1.09$\times$} & \hlg 4.21 & \hlg\textbf{1.31$\times$} \\
          & DFlash (\textbf{16}) & 6.10 & 0.75$\times$ & 7.76 & 1.10$\times$ & 6.41 & 0.82$\times$ & 5.59 & 0.86$\times$ & 3.19 & 0.48$\times$ & 5.81 & 0.80$\times$ \\
          & \hl\method (\textbf{16}) & \hl 5.68 & \hl 1.08$\times$ & \hl 6.86 & \hl 1.49$\times$ & \hl 5.77 & \hl 1.18$\times$ & \hl 4.75 & \hl 1.32$\times$ & \hl 3.01 & \hl 0.97$\times$ & \hl 5.21 & \hl 1.21$\times$ \\
        \midrule
        \multirow{5}{*}{\shortstack{Qwen3.5\\27B}}
          & MTP        & 2.85 & 1.33$\times$ & 2.86 & 1.37$\times$ & 2.83 & 1.32$\times$ & 2.73 & \textbf{1.29$\times$} & 2.53 & \textbf{1.22$\times$} & 2.76 & \textbf{1.31$\times$} \\
          & DFlash (\textbf{8})  & 5.61 & 1.40$\times$ & 5.88 & 1.41$\times$ & 5.63 & 1.16$\times$ & 4.80 & 1.18$\times$ & 3.42 & 0.88$\times$ & 5.07 & 1.21$\times$ \\
          & \hlg\method (\textbf{8})  & \hlg 5.26 & \hlg 1.41$\times$ & \hlg 5.46 & \hlg 1.38$\times$ & \hlg 5.38 & \hlg 1.32$\times$ & \hlg 4.27 & \hlg 1.09$\times$ & \hlg 3.21 & \hlg 0.98$\times$ & \hlg 4.72 & \hlg 1.24$\times$ \\
          & DFlash (\textbf{16}) & 7.17 & 1.14$\times$ & 7.71 & 1.09$\times$ & 7.63 & 1.12$\times$ & 5.58 & 0.78$\times$ & 3.71 & 0.57$\times$ & 6.36 & 0.94$\times$ \\
          & \hl\method (\textbf{16}) & \hl 6.55 & \hl\textbf{1.44$\times$} & \hl 6.93 & \hl\textbf{1.44$\times$} & \hl 7.04 & \hl\textbf{1.38$\times$} & \hl 5.06 & \hl 1.11$\times$ & \hl 3.47 & \hl 0.90$\times$ & \hl 5.81 & \hl 1.25$\times$ \\
        \midrule
        \multirow{5}{*}{\shortstack{Qwen3.5\\35B-A3B}}
          & MTP        & 2.81 & 1.71$\times$ & 2.83 & 1.86$\times$ & 2.79 & 1.77$\times$ & 2.71 & 1.75$\times$ & 2.48 & 1.66$\times$ & 2.72 & 1.75$\times$ \\
          & DFlash (\textbf{8})  & 5.27 & 2.65$\times$ & 5.41 & 2.75$\times$ & 5.29 & 2.39$\times$ & 4.39 & 2.19$\times$ & 3.21 & 1.78$\times$ & 4.71 & 2.35$\times$ \\
          & \hlg\method (\textbf{8})  & \hlg 5.02 & \hlg 2.45$\times$ & \hlg 5.16 & \hlg 2.75$\times$ & \hlg 5.06 & \hlg 2.34$\times$ & \hlg 4.25 & \hlg\textbf{2.24$\times$} & \hlg 3.18 & \hlg\textbf{1.81$\times$} & \hlg 4.53 & \hlg 2.32$\times$ \\
          & DFlash (\textbf{16}) & 6.48 & \textbf{2.78$\times$} & 6.93 & 2.95$\times$ & 6.77 & 2.41$\times$ & 5.16 & 2.06$\times$ & 3.46 & 1.59$\times$ & 5.76 & 2.36$\times$ \\
          & \hl\method (\textbf{16}) & \hl 6.20 & \hl 2.66$\times$ & \hl 6.53 & \hl\textbf{3.04$\times$} & \hl 6.60 & \hl\textbf{2.77$\times$} & \hl 4.98 & \hl 2.21$\times$ & \hl 3.36 & \hl 1.80$\times$ & \hl 5.53 & \hl\textbf{2.50$\times$} \\
        \midrule
        \multirow{5}{*}{\shortstack{Hy3\\295B-A21B}}
          & MTP        & 1.86 & 1.36$\times$ & 1.89 & 1.38$\times$ & 1.87 & 1.36$\times$ & 1.84 & 1.37$\times$ & 1.68 & 1.25$\times$ & 1.83 & 1.34$\times$ \\
          & DFlash (\textbf{8})  & 4.86 & 2.50$\times$ & 4.79 & 2.59$\times$ & 4.78 & 2.42$\times$ & 4.56 & 2.47$\times$ & 2.87 & 1.65$\times$ & 4.37 & 2.33$\times$ \\
          & \hlg\method (\textbf{8})  & \hlg 4.81 & \hlg 2.51$\times$ & \hlg 4.66 & \hlg 2.64$\times$ & \hlg 4.71 & \hlg 2.50$\times$ & \hlg 4.47 & \hlg 2.57$\times$ & \hlg 2.86 & \hlg\textbf{1.77$\times$} & \hlg 4.30 & \hlg 2.40$\times$ \\
          & DFlash (\textbf{16}) & 5.87 & 1.88$\times$ & 5.69 & 2.12$\times$ & 5.84 & 2.02$\times$ & 5.33 & 1.94$\times$ & 3.00 & 1.21$\times$ & 5.15 & 1.83$\times$ \\
          & \hl\method (\textbf{16}) & \hl 5.53 & \hl\textbf{2.61$\times$} & \hl 5.54 & \hl\textbf{2.79$\times$} & \hl 5.63 & \hl\textbf{2.61$\times$} & \hl 5.15 & \hl\textbf{2.64$\times$} & \hl 2.98 & \hl 1.73$\times$ & \hl 4.97 & \hl\textbf{2.48$\times$} \\
        \bottomrule
    \end{tabular}%
    }
\end{table*}

\Cref{tab:main-results} demonstrates that long, fixed-depth verification can yield high MAT, but this does not reliably translate into throughput gains.
For instance, while DFlash (16) generally accepts more tokens per step than DFlash (8), the inflated target-verification batch can cause it to run slower than standard autoregressive decoding on several benchmarks.
This empirically validates our core motivation (\Cref{sec:motivation-runtime-cost}): accepted length alone is an insufficient metric because the runtime verification cost dictates whether computing those extra tokens is actually worthwhile.

\method resolves this trade-off by selectively pruning low-utility suffixes while preserving high-confidence prefixes.
Compared to DFlash (16), \method (16) accelerates throughput in 29 out of 30 model-dataset configurations, elevating the average speedup from $1.26\times$ to $1.65\times$.
These gains are most pronounced where DFlash (16) achieves solid MAT but suffers from poor runtime efficiency, such as Llama-3.1-8B on GSM8K and Qwen3-8B on MT-Bench.
By dynamically trimming the draft, \method retains the high acceptance benefits of long proposals while bypassing their excessive verification costs.

Even under the shorter (8) setting, where there is inherently less room to prune, \method (8) still outperforms its baseline in 25 out of 30 pairs, raising the average speedup from $1.56\times$ to $1.67\times$.
As expected, these improvements are smaller because the 8-token proposal leaves fewer low-utility positions to remove.

\subsection{Scaling with Concurrency}
\label{sec:experiments-concurrency}

\begin{figure}[t]
    \centering
    \includegraphics[width=\textwidth]{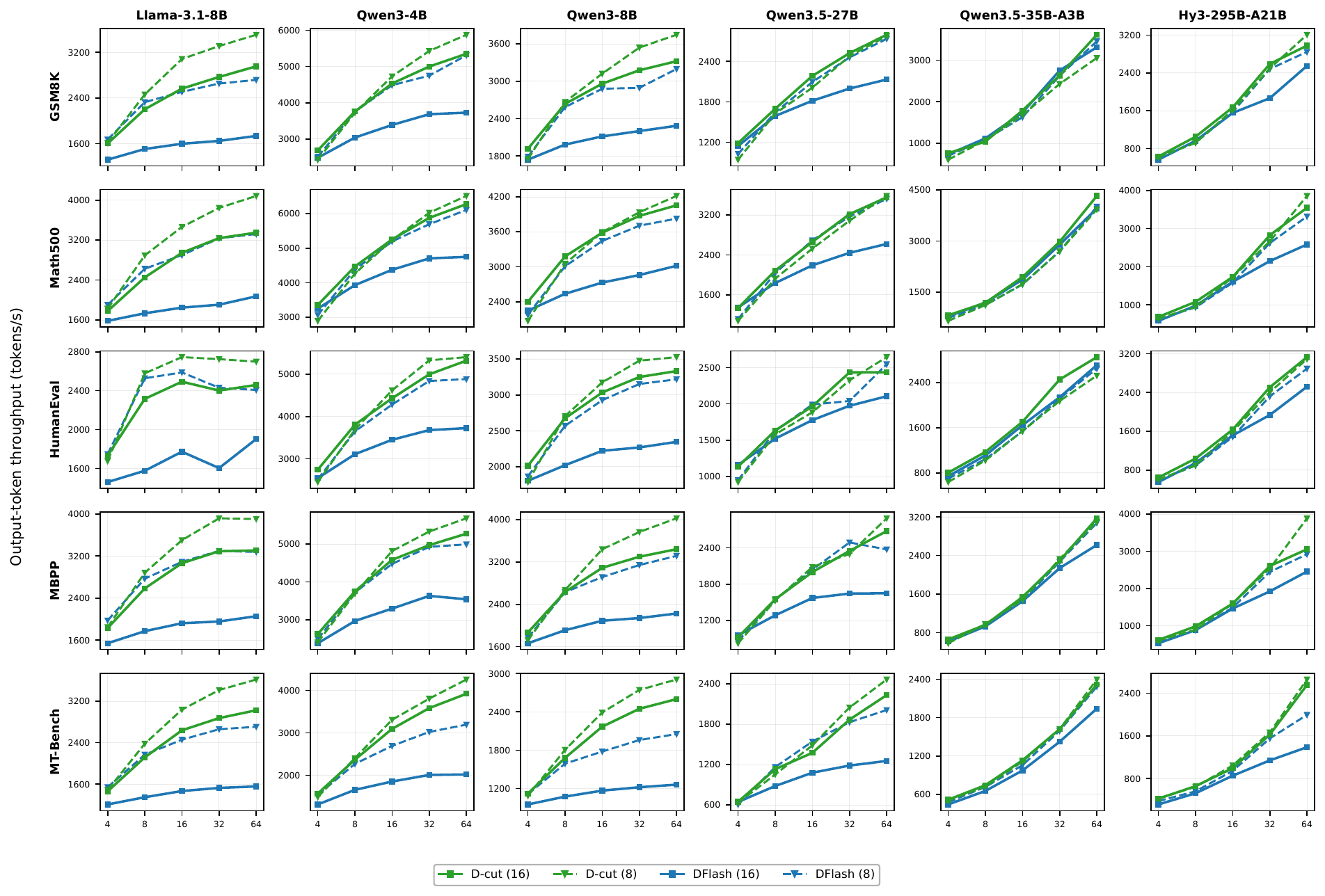}
    \caption{\textbf{Absolute output-token throughput (tokens/s) as concurrency grows from 4 to 64.} Rows are datasets and columns are target models. Green is \method and blue is DFlash; solid lines are the (16) budget and dashed lines the (8) budget. Throughput generally grows with concurrency for both methods. On the dense targets, the long-draft DFlash (16) gains the least at high concurrency, whereas the \method variants sustain higher throughput. The two MoE targets differ: DFlash and \method track closely on Qwen3.5-35B-A3B, while Hy3-295B-A21B benefits more clearly from pruning.}
    \label{fig:concurrency-throughput}
\end{figure}

While \Cref{tab:main-results} fixes concurrency at 32, \Cref{fig:concurrency-throughput} illustrates absolute output-token throughput as concurrency scales from 4 to 64.
Throughput naturally rises with concurrency across methods, but \method and DFlash diverge significantly at the high end.
On dense targets, the long-draft DFlash (16) gains the least from batching, as the high cost of verifying long drafts effectively cancels out the benefits of accepted tokens.
In contrast, \method (16) aggressively prunes low-utility suffixes, sustaining much higher throughput across the high-concurrency regime, with the performance gap widening as load increases.
On the two MoE targets, \method and DFlash track closely on Qwen3.5-35B-A3B, whereas Hy3-295B-A21B benefits clearly from pruning, especially under the longer (16) budget.
One plausible contributor is target-side attention: Qwen3.5-35B-A3B uses a hybrid layout with three Gated DeltaNet layers for every gated full-attention layer, while Hy3-295B-A21B uses GQA with 64 query heads and 8 key-value heads~\citep{qwen35blog,hy3}.
Because model scale and tensor parallelism also differ, this comparison does not isolate the attention effect; \method therefore selects the verification budget from profiled runtime cost rather than from architecture labels.

\subsection{Budget Selection and Runtime Adaptivity}
\label{sec:experiments-ablation}

\begin{table*}[t]
    \centering
    \small
    \setlength{\tabcolsep}{6pt}
    \renewcommand{\arraystretch}{1.15}
    \definecolor{HYLightBlue}{HTML}{A8DFF6}
    \definecolor{HYLightGreen}{HTML}{CDEBD6}
    \caption{\textbf{Throughput ablation on the budget-selection rule (Qwen3-8B).} Output-token throughput (tokens/s) per benchmark at concurrencies 16--64, on H20 and H800. \method (0.25/0.5/0.75) are fixed-ratio variants that verify a global batch budget equal to that fraction of all $B(D{+}1)$ block positions; \method (auto) selects the budget from the profiled runtime cost. The best budget per column is shaded blue (\textbf{bold}) and the runner-up green (\underline{underlined}). On the weaker H20, verification is the bottleneck and aggressive pruning wins; on the compute-rich H800, the best fixed ratio shifts toward retaining more positions, while \method (auto) stays at or near the top on both devices.}
    \label{tab:ablation}
    \resizebox{0.95\textwidth}{!}{%
    \begin{tabular}{ll rrr rrr rrr}
        \toprule
        & & \multicolumn{3}{c}{\textbf{GSM8K}} & \multicolumn{3}{c}{\textbf{HumanEval}} & \multicolumn{3}{c}{\textbf{MT-Bench}} \\
        \cmidrule(lr){3-5}\cmidrule(lr){6-8}\cmidrule(lr){9-11}
        \textbf{Device} & \textbf{Method} & bs=16 & bs=32 & bs=64 & bs=16 & bs=32 & bs=64 & bs=16 & bs=32 & bs=64 \\
        \midrule
        \multirow{5}{*}{H20}
        & DFlash & 2115 & 2200 & 2285 & 2221 & 2268 & 2346 & 1167 & 1220 & 1261 \\
        & \method (0.25) & 2764 & 3014 & 3123 & 2754 & 2972 & 3127 & \cellcolor{HYLightGreen}\underline{2152} & \cellcolor{HYLightGreen}\underline{2377} & \cellcolor{HYLightGreen}\underline{2564} \\
        & \method (0.5) & \cellcolor{HYLightGreen}\underline{2891} & \cellcolor{HYLightGreen}\underline{3114} & \cellcolor{HYLightGreen}\underline{3147} & \cellcolor{HYLightGreen}\underline{3005} & \cellcolor{HYLightGreen}\underline{3179} & \cellcolor{HYLightBlue}\textbf{3351} & 1729 & 1856 & 1941 \\
        & \method (0.75) & 2427 & 2618 & 2630 & 2463 & 2716 & 2796 & 1355 & 1442 & 1514 \\
        & \method (auto) & \cellcolor{HYLightBlue}\textbf{2961} & \cellcolor{HYLightBlue}\textbf{3175} & \cellcolor{HYLightBlue}\textbf{3320} & \cellcolor{HYLightBlue}\textbf{3035} & \cellcolor{HYLightBlue}\textbf{3250} & \cellcolor{HYLightGreen}\underline{3333} & \cellcolor{HYLightBlue}\textbf{2169} & \cellcolor{HYLightBlue}\textbf{2449} & \cellcolor{HYLightBlue}\textbf{2600} \\
        \midrule
        \multirow{5}{*}{H800}
        & DFlash & 5963 & 7912 & 7908 & 6492 & 8167 & 8076 & 3159 & 4147 & 4245 \\
        & \method (0.25) & 4814 & 5479 & 6143 & 4735 & 7251 & 9520 & 3387 & \cellcolor{HYLightBlue}\textbf{5495} & \cellcolor{HYLightBlue}\textbf{7126} \\
        & \method (0.5) & \cellcolor{HYLightBlue}\textbf{6615} & \cellcolor{HYLightBlue}\textbf{9780} & \cellcolor{HYLightGreen}\underline{10824} & 6799 & \cellcolor{HYLightGreen}\underline{9527} & \cellcolor{HYLightGreen}\underline{10831} & \cellcolor{HYLightBlue}\textbf{3562} & 5185 & 6041 \\
        & \method (0.75) & \cellcolor{HYLightGreen}\underline{6541} & 8661 & 9551 & \cellcolor{HYLightGreen}\underline{6809} & 8962 & 9827 & 3404 & 4594 & 5049 \\
        & \method (auto) & 6391 & \cellcolor{HYLightGreen}\underline{9663} & \cellcolor{HYLightBlue}\textbf{11194} & \cellcolor{HYLightBlue}\textbf{6823} & \cellcolor{HYLightBlue}\textbf{9737} & \cellcolor{HYLightBlue}\textbf{11347} & \cellcolor{HYLightGreen}\underline{3469} & \cellcolor{HYLightGreen}\underline{5253} & \cellcolor{HYLightGreen}\underline{6894} \\
        \bottomrule
    \end{tabular}}
\end{table*}

To isolate the impact of our budget-selection rule, we evaluate Qwen3-8B with the block-16 drafter across five configurations: unpruned DFlash, three fixed-ratio variants of \method (using a global batch budget of $0.25$, $0.50$, or $0.75$ of all block positions), and the fully runtime-adaptive \method (auto).
\Cref{tab:ablation} reports the resulting throughput across concurrencies on both H20 and H800 GPUs.

The results reveal that the optimal pruning ratio is highly dynamic, varying by both benchmark and hardware.
Across the contended regime in \Cref{tab:ablation}, pruning improves over the unpruned DFlash baseline in many settings, yet no single fixed ratio is best everywhere.
On the compute-constrained H20, verification is a severe bottleneck. Here, MT-Bench heavily rewards aggressive pruning (\method (0.25) reaches $2{,}564$ tokens/s at concurrency 64), whereas GSM8K and HumanEval favor a more moderate $0.50$ ratio.
Conversely, on the compute-rich H800, verifying extra tokens is significantly cheaper. The larger $0.50$ budget becomes optimal for GSM8K and HumanEval, rendering the $0.25$ ratio overly aggressive and wasteful, while MT-Bench still benefits from aggressive pruning at the high-concurrency end because its accepted prefixes are shorter.
Clearly, no single static ratio universally excels.

\method (auto) tracks this shifting optimum without requiring manual, per-deployment tuning.
By reading the profiled cost table at startup and dynamically selecting the budget, it stays near the best fixed ratio across devices and benchmarks.
For instance, at concurrency 64 on H800 (GSM8K), \method (auto) achieves $11{,}194$ tokens/s, outperforming both the best fixed ratio ($10{,}824$) and DFlash ($7{,}908$).
Averaged across the three benchmarks and concurrencies 16 to 64, \method (auto) accelerates DFlash by $1.58\times$ on the H20 and $1.25\times$ on the H800, matching the best fixed configuration at the aggregate level without requiring the ratio to be chosen in advance.

\subsection{Robustness across Sampling Regimes}
\label{sec:experiments-temperature}

\begin{figure}[t]
    \centering
    \includegraphics[width=\textwidth]{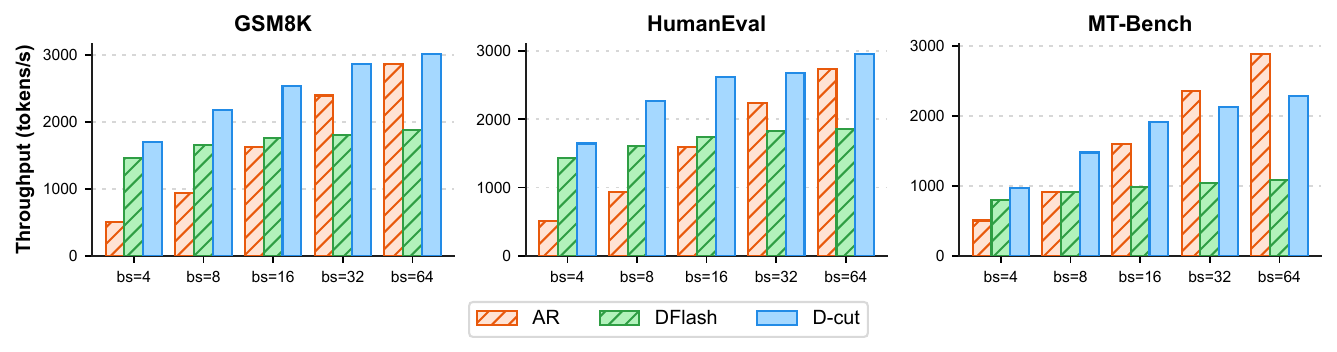}
    \caption{\textbf{Robustness under temperature-1 target sampling (Qwen3-8B, H20).} Output-token throughput (tokens/s) as batch size grows from 4 to 64, with greedy draft proposals and target-only verification. Target-side stochastic sampling shortens accepted drafts, so the long-draft DFlash gains little and is overtaken by autoregressive (AR) decoding at high concurrency. \method (auto) prunes the now even less useful suffix and remains faster than DFlash across the sweep, sustaining a $1.15\times$ average speedup over AR.}
    \label{fig:temperature}
\end{figure}

The results so far evaluate greedy target decoding.
\Cref{fig:temperature} repeats the Qwen3-8B H20 sweep with temperature-1 target sampling while retaining greedy draft proposals and target-only verification.
In this setting, target-side stochastic sampling reduces draft utilization relative to greedy target decoding, which severely penalizes long, fixed-depth proposals.
Averaged over the three benchmarks and concurrencies 16 to 64, DFlash's speedup collapses to $0.68\times$ over autoregressive decoding, falling strictly below the baseline at high concurrency.
By aggressively pruning the now even less useful suffixes, \method (auto) sustains a $1.15\times$ speedup over the autoregressive baseline, translating to a $1.68\times$ relative gain over DFlash.
This relative advantage actually exceeds the $1.58\times$ gain observed under greedy decoding (\Cref{sec:experiments-ablation}), demonstrating that harsher sampling regimes make adaptive pruning more critical, not less.

\section{Related Work}
\label{sec:related_work}

Speculative decoding accelerates autoregressive generation through a lossless \emph{draft-then-verify} paradigm~\citep{sps1, sps2}, and a long line of work improves the drafter, from token trees~\citep{specinfer, sequoia, opt-tree} and multi-head prediction~\citep{medusa, mtp} to feature-reusing autoregressive drafters~\citep{eagle, eagle2, eagle3, glide} and, most recently, block-parallel diffusion drafters such as DFlash that emit a whole draft block in one forward pass~\citep{dflash}.
Most prior adaptive methods change the draft length or tree online to avoid wasting verification on low-confidence candidates~\citep{specdec++, disco, talon, specbranch, dyspec}, but their control objective remains each request's local draft structure.
Closest to our setting, recent work shows speculative speedups shrink under high concurrency as batched verification turns compute-bound~\citep{illusion, echo}; PEARL overlaps drafting and verification with adaptive draft length~\citep{pearl}, and ECHO casts high-concurrency speculation as budgeted tree scheduling under a global verification cap~\citep{echo}.
\method shares the view that verification compute is the scarce resource in batched serving, but targets a different drafter class: ECHO allocates its global verification budget over tree-structured or autoregressive drafters such as EAGLE and MTP~\citep{mtp,li2026breaking}, scheduling tree depth and width, whereas DFlash emits a single linear block per request in one parallel pass. For such block-parallel drafts, the key decision is the per-request verification depth, so \method prunes linear draft blocks across requests and sets the depth from a profiled runtime cost model, making the allocation adaptive to both batch confidence and deployment hardware.
DSpark shares \method's view of cross-request pruning and runtime-adaptive verification~\citep{dspark}.
It uses a semi-autoregressive drafter and a trained confidence head, whereas \method reuses confidence from an existing block-parallel drafter without additional training.
A closely related approach, Graft, also prunes low-confidence draft positions and reuses the verification budget freed by pruning, but it refills that budget with retrieved candidates within each request to restore accepted length~\citep{graft}; \method instead redistributes the freed budget \emph{across} requests under a profiled runtime cost model, so the two mechanisms are complementary.
A full discussion is given in \Cref{sec:appendix-related}.

\section{Conclusion and Limitations}
\label{sec:conclusion}

As large language models are deployed at scale, the focus of speculative decoding must shift from simply maximizing single-request acceptance lengths to managing batched verification efficiency.
We demonstrated that block-parallel drafters, while highly effective in isolation, can degrade throughput at high concurrency because their long, fixed-depth drafts waste scarce target-model compute on low-yield tokens.

\method addresses this by treating target verification as a strictly budgeted resource.
By formally modeling the speedup as a ratio of confidence-estimated benefit (MAT) to hardware-profiled runtime cost, \method dynamically prunes low-utility draft suffixes across the batch.
This ensures that the engine attempts deeper drafts only when the specific hardware and concurrency level make it profitable to do so.
This acceleration is achieved entirely at the scheduling level, strictly preserving the target model's original output distribution.
Evaluations across dense and MoE architectures show that \method keeps long-draft configurations from falling below autoregressive baselines at high load, lifting average speedups from $1.26\times$ to $1.65\times$.

A practical limitation of \method lies in its system integration.
Its per-step pruning produces a verification batch whose shape varies across steps, which does not fit cleanly into the cpu-gpu overlap (Spec-V2) and the full-plus-piecewise CUDA-graph capture of current inference engines, both of which assume a static per-step shape.
Realizing the full benefit of dynamic verification therefore requires non-trivial engineering, and tighter co-design with the serving framework is a direction for future work.

\section*{Acknowledgments}

We thank Xiang Li, Yepeng Weng, Xin Cheng, Jingzhou Chen, Jincheng Xie, and Jiale Fu for helpful discussions on the losslessness of \method.

\newpage

\bibliographystyle{citation}
\bibliography{citation}

\newpage
\appendix
\section{Algorithm and Analysis}
\label{sec:appendix-algorithm}

\Cref{alg:dcut} summarizes how \method executes inside a single DFlash decoding step: the selector runs after the drafter samples a block and before the target model verifies it, ranking block positions across requests, choosing a runtime-aware budget, and packing the retained prefixes for verification.

\begin{algorithm}[h]
\caption{\method at one decoding step.}
\label{alg:dcut}
\begin{algorithmic}[1]
\Require Active requests $\mathcal{B}$, draft depth $D$, ratio buckets $\mathcal{R}$, runtime cost table $C$
\Ensure Packed target-verification batch
\State Run the DFlash drafter to obtain draft tokens $z_{i,1:D}$ and confidences $c_{i,1:D}$ for each request $i \in \mathcal{B}$
\State Set $s_{i,0} \gets 1$ and compute prefix scores $s_{i,k} \gets \prod_{t=1}^{k} c_{i,t}$ for all block positions $(i,k)$, $k=0,\ldots,D$
\ForAll{$\rho \in \mathcal{R}$}
    \State Compute $K_{\rho}(B)$ using \Cref{eq:dcut-budget}
    \State Let $\mathcal{S}_{K_{\rho}(B)}$ be the top-$K_{\rho}(B)$ positions ranked by $s_{i,k}$ (ties toward smaller $k$)
    \State Score $\rho$ by the utility-cost objective in \Cref{eq:dcut-objective}
\EndFor
\State Select $\rho^\star$ with the largest objective value
\ForAll{request $i \in \mathcal{B}$}
    \State $n_i \gets \max(\{k:(i,k)\in \mathcal{S}_{K_{\rho^\star}(B)}\}\cup\{0\})$
    \State Keep the block prefix $z_{i,0:n_i}$ and prune the remaining draft suffix
\EndFor
\State Pack all retained block prefixes for target verification
\end{algorithmic}
\end{algorithm}

\paragraph{Selector complexity.}
Each step the selector accumulates log-confidences to obtain the prefix scores of all $B(D{+}1)$ block positions (\Cref{eq:dcut-prefix-score}), which is $O(B(D{+}1))$.
It then flattens the positions and extracts the top-$K_{\rho}(B)$ set; ranking the $B(D{+}1)$ scores costs $O\!\bigl(B(D{+}1)\log B(D{+}1)\bigr)$, and scoring the objective in \Cref{eq:dcut-objective} for each of the $|\mathcal{R}|$ ratios is an $O(B(D{+}1))$ prefix-sum sweep over the ranked list.
The per-step cost is thus $O\!\bigl(B(D{+}1)\log B(D{+}1)\bigr)$ and independent of the target model size, whereas verification cost grows with both the model size and the retained budget $K$.
This asymmetry is why the measured selector cost stays a small fraction of the step (\Cref{fig:overhead}).

\paragraph{Retained positions form a contiguous prefix.}
The selector ranks positions by the prefix-product score $s_{i,k}=\prod_{t=1}^{k}c_{i,t}$ (\Cref{eq:dcut-prefix-score}), which is non-increasing in $k$ because $s_{i,k}=s_{i,k-1}\,c_{i,k}$ with $c_{i,k}\in[0,1]$.
For a fixed request $i$ this gives $s_{i,j}\ge s_{i,k}$ whenever $j<k$.
Suppose the global top-$K$ set $\mathcal{S}_{K}$ contains a position $(i,k)$; then every shallower position $(i,j)$ with $j<k$ has a score at least as large and is therefore also in $\mathcal{S}_{K}$.
Hence each request's retained positions are exactly $\{0,1,\dots,n_i\}$ with $n_i=\max\{k:(i,k)\in\mathcal{S}_{K}\}$, a contiguous prefix of the draft block rather than a scattered subset.
This is the property that lets \method hand the verifier a valid shortened block and reuse the engine's piecewise CUDA-graph shapes: cross-request pruning only ever removes a suffix from a request, never an interior position.

\paragraph{Correctness.}
\label{sec:appendix-execution}
In the evaluated setup, draft proposals are deterministic, and the temperature-1 study changes only target-side sampling and uses target-only verification.
\method selects the keep depths before target samples are drawn and leaves the target conditional distributions unchanged, so truncating each block to a prefix changes only how many target positions are evaluated in parallel, not the distribution of committed tokens.
Progress never stalls either, since the budget is clamped to $K_{\rho}(B)\ge B$ (\Cref{eq:dcut-budget}) so every request keeps at least its bonus token $z_{i,0}$ and advances by at least one token per step.

\section{Causality under Rejection Sampling}
\label{sec:appendix-rs}

\paragraph{Selection bias from inclusive confidence.}
We now consider an extension in which each request exposes the per-position conditional proposal distributions $q_{i,k}(\cdot\mid z_{i,<k})$ required by standard rejection sampling (RS), and the target distribution $p$ verifies proposals with independent sampling randomness across requests~\citep{sps1, sps2}.
In this setting, directly applying the original score $s_{i,k}=\prod_{t=1}^{k}q_{i,t}(z_{i,t})$ can change the output distribution: whether position $k$ is retained depends on the sampled proposal $z_{i,k}$ itself, violating the non-anticipating condition required by RS~\citep{dspark}.
For a one-position example, let the vocabulary be $\{A,B\}$, with $p(A)=p(B)=0.5$ and $q(A)=0.9$, $q(B)=0.1$, and retain the proposal only when $q(z)>0.5$.
When $A$ is proposed, RS accepts it with probability $0.5/0.9=5/9$ and otherwise emits $B$ from the residual distribution; when $B$ is proposed, the selector skips the proposal and samples directly from $p$.
The resulting probability of emitting $A$ is therefore
\begin{equation}
    0.9\cdot\frac{5}{9}+0.1\cdot 0.5=0.55\neq p(A),
\end{equation}
showing that leaving the RS accept/reject rule unchanged is insufficient when the retention decision depends on the current proposal.

\paragraph{Shifted confidence with causal budget selection.}
To make the scheduling decision causal, we shift each draft-position score by one position:
\begin{equation}
    \tilde{s}_{i,0}=1,\qquad
    \tilde{s}_{i,k}=\prod_{t=1}^{k-1}q_{i,t}(z_{i,t}),\quad k=1,\ldots,D.
    \label{eq:dcut-shifted-score}
\end{equation}
The admission score for position $k$ then depends only on proposals before $k$.
The shift alone is not sufficient if the selector retrospectively chooses the budget from all scores in the current block: $z_{i,k}$ affects $\tilde{s}_{i,k+1}$ and can thereby change the selected budget, which may in turn change whether position $k$ is retained.
Our preferred solution is a causal early-stopping scan, following the non-anticipating construction in DSpark~\citep{dspark}.
The selector maintains one frontier position per request and repeatedly considers the frontier with the largest shifted score.
It evaluates the utility-cost objective at the resulting verification size; if the objective improves, the position is irrevocably admitted and the next position from that request is exposed, otherwise the scan stops.
This determines $K$ from the current block without consulting a descendant score before deciding on its ancestor.
Unlike a retrospective global argmax, early stopping may miss a later optimum when the utility-cost curve is not unimodal.
Unimodality affects budget optimality only, not the distributional correctness established below.
When token-level stopping does not fit an asynchronous or bucketed execution path, an alternative is to estimate $K$ from confidence statistics of previous decoding steps, following the history-based scheduling used in DSpark implementations~\citep{dspark}.
The resulting history-determined $K$ is fixed before the current block is observed, after which the selector applies global top-$K$ allocation to the shifted scores.
Both variants resolve ties deterministically by shallower depth and then request index, independently of proposal values.
The verifier still uses the original $p$ and $q$ distributions in the RS correction; shifted confidence affects scheduling only.

\paragraph{Distributional correctness.}
Condition on the committed history and on the independent proposal randomness of other requests.
For position $(i,k)$, the shifted score depends only on $z_{i,<k}$, while all deeper scores from the same request are no larger than $\tilde{s}_{i,k}$.
Under causal early stopping, the decision to admit $(i,k)$ is made before any descendant score from request $i$ is exposed; under history-based scheduling, $K$ is fixed before the current proposals are drawn and every descendant is ranked after $(i,k)$.
In either case, changing $z_{i,k}$ or any later proposal cannot change whether $(i,k)$ is retained.
Thus the retention event is independent of $z_{i,k}$ conditional on $z_{i,<k}$, and the proposal at position $k$ still follows $q_{i,k}$ after conditioning on retention.
Standard RS then accepts with probability $\min(1,p_{i,k}(z_{i,k})/q_{i,k}(z_{i,k}))$ and, upon rejection, samples from the residual distribution proportional to $(p_{i,k}-q_{i,k})_{+}$; if the retained block is fully accepted, the bonus token is sampled directly from the next target distribution.
The token committed at the first unresolved position therefore follows $p_{i,k}$.
The shared top-$K$ decision can couple how far different requests advance, but it does not change their conditional token distributions; with independent sampling randomness across requests, induction over positions and decoding steps recovers the target joint output distribution.

\section{System Implementation}
\label{sec:appendix-system}

\paragraph{Packing and CUDA-graph compatibility.}
The retained prefixes have unequal lengths $n_i{+}1$ across requests.
\method concatenates them into a single contiguous verification batch of $K_{\rho^\star}(B)$ positions, together with the block and position metadata the target attention needs to score each retained position against its accepted prefix.
Because the budget is restricted to the ratio buckets $\mathcal{R}$ (\Cref{eq:dcut-budget}), the packed batch takes one of only a few total lengths, each of which overlaps the piecewise CUDA graphs the engine already captures for ordinary decoding.
Pruning therefore changes how many positions are packed, not the set of captured shapes, so it adds no extra graphs and no additional graph memory.
A freely varying $K$, by contrast, would produce a new shape almost every step and force the engine to capture and store many more graphs.

\paragraph{Cost-table profiling protocol.}
\method builds the cost table $C(B,\rho)$ once at server startup, before the server accepts traffic.
For each batch size $B$ that the engine captures a piecewise CUDA graph for, and each ratio $\rho \in \mathcal{R}$, we run dummy speculative steps that exercise the full pipeline, drafting, selection, packing, and target verification of the $K_{\rho}(B)$ retained positions, under the same model weights, parallelism, compilation mode, and CUDA-graph configuration as serving, and record the median per-step latency over a small number of repetitions.
Each entry is therefore an end-to-end $T_{\mathrm{spec}}$ measurement rather than a verification-only cost, consistent with \Cref{eq:dcut-objective}.
At decode time the selector reads $C(B,\rho)$ at the current batch size $B$ in $O(1)$; batch sizes that fall between captured points reuse the nearest captured entry.
Because profiling touches only dummy tensors and runs before serving, it adds a one-time startup cost and no per-request latency.
\Cref{tab:cost-table-case-study} shows concrete tables from the Qwen3-8B H20 and H800 runs.
The cost is nearly flat at $B=1$, where verification is underutilized, but grows sharply with the retained ratio at larger batches, especially on the compute-constrained H20.
At $B=64$, verifying the full DFlash block costs $133.91$\,ms on H20 and $30.39$\,ms on H800, while the $25\%$ bucket costs $55.16$\,ms and $14.46$\,ms, respectively.
This changing slope across both batch size and hardware is the reason \method reads the profiled table at runtime instead of treating each extra draft position as having a constant cost.

\begin{table}[h]
    \centering
    \scriptsize
    \setlength{\tabcolsep}{3.4pt}
    \renewcommand{\arraystretch}{1.15}
    \caption{\textbf{Example startup cost table $C(B,\rho)$ for Qwen3-8B on H20 and H800.} Each entry is the profiled end-to-end speculative-step latency in milliseconds for a retained-position budget $K_\rho(B)=\rho B(D+1)$, with $D=15$ and ratio buckets $\mathcal{R}=\{25\%,50\%,75\%,100\%\}$.}
    \label{tab:cost-table-case-study}
    \begin{tabular}{rcccccccc}
        \toprule
        \multirow{2}{*}{\textbf{$B$}} &
        \multicolumn{4}{c}{\textbf{H20}} &
        \multicolumn{4}{c}{\textbf{H800}} \\
        \cmidrule(lr){2-5}\cmidrule(lr){6-9}
        & $25\%$ & $50\%$ & $75\%$ & $100\%$
        & $25\%$ & $50\%$ & $75\%$ & $100\%$ \\
        \midrule
         1 &  7.180 &  7.157 &   7.310 &   7.353 &  7.995 &  8.008 &  8.081 &  8.101 \\
         8 & 10.364 & 12.775 &  16.454 &  19.663 &  8.363 &  8.457 &  8.657 &  8.628 \\
        16 & 15.991 & 22.785 &  30.003 &  35.869 &  9.008 &  9.079 &  9.700 & 10.208 \\
        32 & 29.903 & 43.000 &  56.611 &  70.066 & 10.408 & 11.554 & 13.793 & 15.926 \\
        64 & 55.162 & 82.287 & 107.874 & 133.914 & 14.464 & 18.904 & 24.317 & 30.388 \\
        \bottomrule
    \end{tabular}
\end{table}

\begin{figure}[h]
    \centering
    \includegraphics[width=\linewidth]{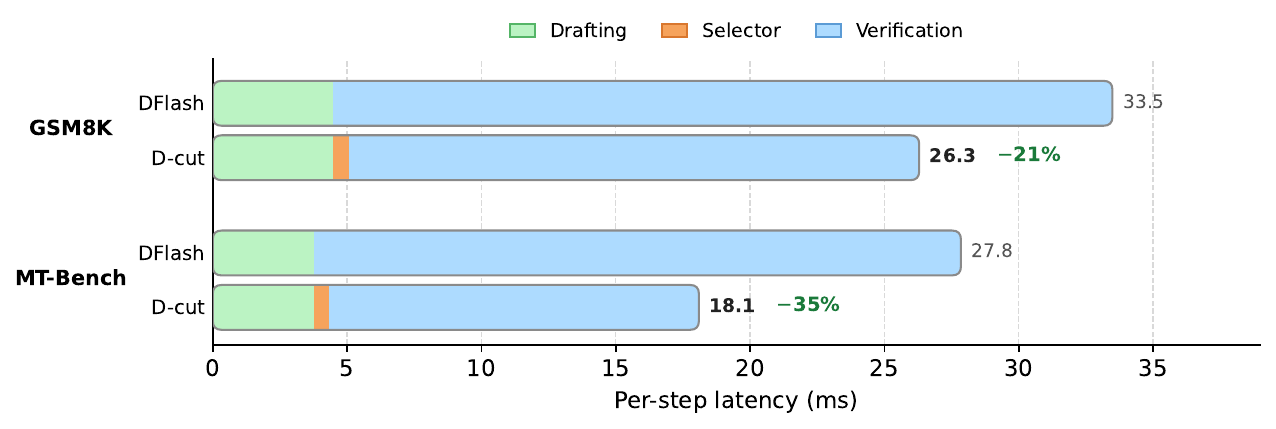}
    \caption{\textbf{Per-step latency at concurrency 64 (Qwen3-8B, H800).} Each bar splits the decoding step, in execution order, into drafting (green), the \method selector (orange), and verification (blue). The drafter is shared, so the drafting bars are aligned; \method's extra prefix-packing cost is folded into verification, which also covers sampling and bookkeeping. Pruning shrinks verification, the dominant cost, and cuts the full step by $21\%$ and $35\%$ on GSM8K and MT-Bench, while the selector adds only $0.56$--$0.58$\,ms, about $2$--$3\%$ of the step.}
    \label{fig:overhead}
\end{figure}

\paragraph{Selector overhead and where the time goes.}
\method adds one routine on top of DFlash: a global top-$K$ ranking of the per-request draft confidences each step, after a one-time cost-table profiling at startup.
\Cref{fig:overhead} splits the per-step latency at concurrency 64, in execution order, into drafting, the \method selector, and verification; the drafter is shared, so the drafting bars are aligned and \method's extra prefix-packing cost is folded into verification (which also covers sampling and bookkeeping).
The selector adds only $0.55$--$0.58$\,ms per step (p95 $\le 0.62$\,ms), which is $2.2$--$3.4\%$ of the full step.
In return, pruning shortens verification, the dominant term, by $23.7$--$38.3\%$ and the full step by $21\%$ and $35\%$ on GSM8K and MT-Bench.
The selector cost is therefore small against the verification it removes, and the net effect is a shorter step.

\section{Experimental Details}
\label{sec:appendix-details}

\paragraph{Draft models and checkpoints.}
\Cref{tab:draft-models} lists the draft model used for each baseline and target. The DFlash drafters are evaluated at verification block sizes of $8$ and $16$.

\begin{table}[h]
    \centering
    \small
    \setlength{\tabcolsep}{6pt}
    \renewcommand{\arraystretch}{1.15}
    \caption{\textbf{Draft models used for each baseline.} DFlash uses the official releases from \url{https://github.com/z-lab/dflash} for the public targets and the corresponding Hy3 checkpoint for Hy3-295B-A21B; EAGLE-3 uses the public AngelSlim heads; MTP reuses the multi-token-prediction head shipped with each target under its default configuration.}
    \label{tab:draft-models}
    \begin{tabular}{lll}
        \toprule
        \textbf{Method} & \textbf{Target} & \textbf{Draft checkpoint} \\
        \midrule
        \multirow{6}{*}{DFlash}
            & Llama-3.1-8B     & \texttt{z-lab/LLaMA3.1-8B-Instruct-DFlash-UltraChat} \\
            & Qwen3-4B         & \texttt{z-lab/Qwen3-4B-DFlash-b16} \\
            & Qwen3-8B         & \texttt{z-lab/Qwen3-8B-DFlash-b16} \\
            & Qwen3.5-27B      & \texttt{z-lab/Qwen3.5-27B-DFlash} \\
            & Qwen3.5-35B-A3B  & \texttt{z-lab/Qwen3.5-35B-A3B-DFlash} \\
            & Hy3-295B-A21B     & \texttt{AngelSlim/Hy3-DFlash-b16} \\
        \midrule
        \multirow{2}{*}{EAGLE-3}
            & Qwen3-4B         & \texttt{AngelSlim/Qwen3-4B\_eagle3} \\
            & Qwen3-8B         & \texttt{AngelSlim/Qwen3-8B\_eagle3} \\
        \midrule
        \multirow{3}{*}{MTP}
            & Qwen3.5-27B      & built-in MTP head \\
            & Qwen3.5-35B-A3B  & built-in MTP head \\
            & Hy3-295B-A21B     & built-in MTP head \\
        \bottomrule
    \end{tabular}
\end{table}

\paragraph{Serving configuration.}
All experiments run on a single node of 8$\times$ NVIDIA H20 GPUs, except the cross-hardware study in \Cref{sec:experiments-ablation}, which uses 8$\times$ NVIDIA H800 GPUs.
The 8B-scale dense models run with tensor parallelism of 1, Qwen3.5-27B and Qwen3.5-35B-A3B use tensor parallelism of 4, and Hy3-295B-A21B uses tensor parallelism of 8.
We build on vLLM and serve each model through its OpenAI-compatible chat endpoint, capping the running batch at 64 sequences (\texttt{max-num-seqs}) with up to 32{,}768 batched tokens per step, a maximum context length of 8{,}192, and 92\% GPU memory utilization.
We use the FlashAttention backend with piecewise CUDA-graph capture, and disable prefix caching and asynchronous scheduling to keep the measured throughput tied to the verification cost of each decoding step.

\paragraph{Benchmarks and decoding.}
For each benchmark, we use a fixed 256-prompt manifest and a maximum output length of 2{,}048 tokens, with greedy decoding (temperature 0); for the Qwen3 and Qwen3.5 models we disable the thinking mode.
We throttle requests to the target concurrency level and warm up each serving configuration before recording throughput, so that the server operates at the intended batch size.

\paragraph{Hyperparameters.}
\Cref{tab:hparams} collects the default configuration, so that the per-section text can stay focused on the variables it sweeps or overrides.

\begin{table}[h]
    \centering
    \small
    \setlength{\tabcolsep}{8pt}
    \renewcommand{\arraystretch}{1.15}
    \caption{\textbf{Default configuration used across experiments.} A few values are overridden by specific studies and stated in the corresponding section, such as the concurrency level and the decoding temperature.}
    \label{tab:hparams}
    \begin{tabular}{ll}
        \toprule
        \textbf{Parameter} & \textbf{Value} \\
        \midrule
        Draft length $D$ & $15$ (verification block of $16$ positions) \\
        Keep ratios $\mathcal{R}$ & $\{0.25,\ 0.50,\ 0.75,\ 1.00\}$ \\
        Max running batch (\texttt{max-num-seqs}) & $64$ \\
        Max batched tokens per step & $32{,}768$ \\
        Max context length & $8{,}192$ \\
        GPU memory utilization & $92\%$ \\
        Tensor parallelism & $1$ (dense 8B); $4$ (Qwen3.5-27B, 35B-A3B); $8$ (Hy3) \\
        Attention backend & FlashAttention \\
        CUDA graph & piecewise capture \\
        Prefix caching / async scheduling & disabled \\
        Decoding temperature & $0$ (greedy); $1$ in \Cref{sec:experiments-temperature} \\
        Thinking mode & off \\
        Prompts per benchmark & $256$ \\
        Max output length & $2{,}048$ \\
        Concurrency sweep & $\{4,\ 8,\ 16,\ 32,\ 64\}$ \\
        \bottomrule
    \end{tabular}
\end{table}

\begin{figure}[h]
    \centering
    \includegraphics[width=0.92\linewidth]{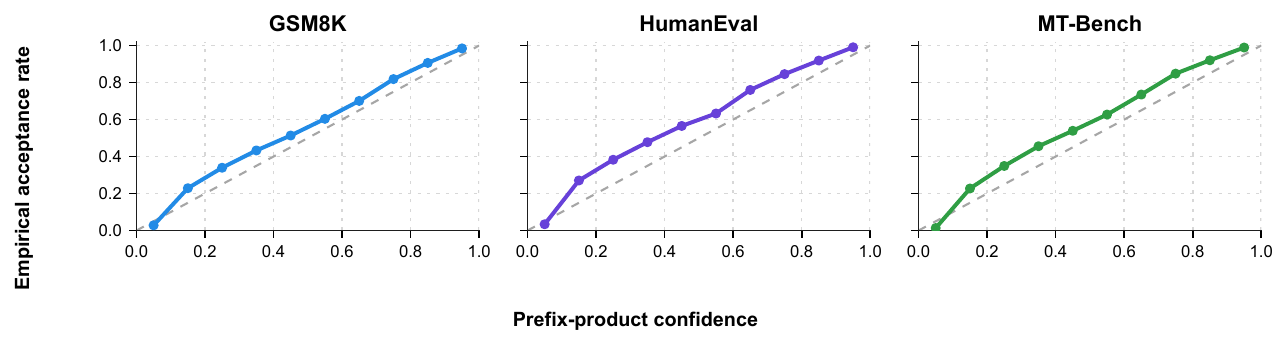}
    \caption{\textbf{Draft confidence versus verifier acceptance.} Token-level acceptance rate after binning DFlash-B16 draft positions by the prefix-product confidence $s_{i,k}$ on Qwen3-8B/H800, using full verification to avoid pruning-induced bias. The dashed gray line marks $y=x$ as a calibration reference. GSM8K, HumanEval, and MT-Bench all show a monotonic confidence-acceptance relationship, confirming that the same confidence used by \method is a strong ranking signal for accepted draft positions.}
    \label{fig:confidence-acceptance}
\end{figure}

\paragraph{Draft confidence predicts token acceptance.}
\label{sec:appendix-confidence}
\Cref{fig:confidence-acceptance} validates the confidence signal used by the selector without the selection bias introduced by \method's pruning: we run DFlash-B16 with full verification on Qwen3-8B/H800, collect every draft position on GSM8K, HumanEval, and MT-Bench at concurrency 32 and 64, and label a position as accepted if it lies inside the verifier's accepted prefix.
The prefix-product confidence $s_{i,k}$ is strongly monotonic with empirical acceptance across all three datasets: Spearman correlation is $0.748$ on GSM8K, $0.770$ on HumanEval, and $0.579$ on MT-Bench, while AUROC is $0.957$, $0.962$, and $0.972$, respectively.
The goal of this signal is ranking rather than probability calibration, and the monotonic curves show that high-confidence positions are precisely the ones most worth preserving under a shared verification budget.

\section{Extended Related Work}
\label{sec:appendix-related}

\paragraph{Speculative decoding and candidate construction.}
Speculative decoding accelerates autoregressive generation through a lossless \emph{draft-then-verify} paradigm, where a lightweight draft process proposes candidate tokens and the target model verifies them in parallel~\citep{sps1, sps2}.
Early methods use independent small models as drafters, while later methods improve candidate quality or diversity through tree verification and multi-head prediction.
SpecInfer organizes candidates as token trees~\citep{specinfer}, Sequoia and OPT-Tree optimize tree structures under token or hardware budgets~\citep{sequoia, opt-tree}, and Medusa adds multiple decoding heads to the target model~\citep{medusa}.
Another branch reduces drafter-target mismatch by reusing target-model information: EAGLE predicts future hidden states from target representations~\citep{eagle}, EAGLE-2/3 improve this line with dynamic trees and multi-layer features~\citep{eagle2, eagle3}, GLIDE reuses the target KV cache~\citep{glide, kvshot}, and multi-token prediction heads provide low-overhead self-drafting candidates~\citep{mtp}.
DFlash further replaces autoregressive drafting with a block-parallel diffusion drafter conditioned on target features, generating an entire draft block in one forward pass~\citep{dflash}.
These works mainly improve how candidates are generated; \method instead focuses on which generated candidates should be verified in a batched step.

\paragraph{Static and dynamic verification budgets.}
Many speculative decoding methods implicitly use a fixed per-request draft shape, such as a fixed chain length or a fixed token tree~\citep{logitspec}.
Such static designs are simple and efficient, but they can waste verification compute on low-utility candidates when the drafter is uncertain.
Adaptive methods address this by changing the draft length or tree online.
SpecDec++ and DISCO adjust candidate length based on confidence or recent acceptance behavior~\citep{specdec++, disco}, while TALON, AdaEAGLE, and DySpec adapt draft-tree structures to the current context~\citep{talon, adaeagle, dyspec}.
Most prior adaptive methods reduce waste within a request, but their control objective is still centered on each request's local draft structure.
In contrast, \method treats the batch as sharing a verification budget and reallocates this budget across requests according to relative draft confidence.

\paragraph{Speculative serving under high concurrency.}
Recent studies show that speculative decoding speedups can diminish under high concurrency, because batched verification must process $\text{bs} \times D$ draft tokens and can move the target model from a memory-bound regime to a compute-bound one~\citep{illusion, echo}.
PEARL addresses serving inefficiency by overlapping drafting and verification with adaptive draft length control~\citep{pearl}.
TETRIS studies cross-request allocation under a fixed verification capacity for autoregressive drafters~\citep{tetris}.
DSpark shares \method's view of cross-request pruning and runtime-adaptive verification~\citep{dspark}.
It uses a semi-autoregressive drafter and a trained confidence head, whereas \method reuses confidence from an existing block-parallel drafter without additional training.
Most closely related to our setting, ECHO formulates high-concurrency speculative decoding as a budgeted scheduling problem and uses sparse confidence gating to coordinate tree depth and width under a global verification cap~\citep{echo}.
\method shares the view that verification compute is the scarce resource in batched serving, but targets a different drafter class: ECHO is designed for tree-structured or autoregressive drafters such as EAGLE and MTP, where the scheduler controls tree depth and width under a global verification cap. DFlash-like block-parallel drafters instead emit a single linear block per request, where the key decision is the verification depth retained for each request.
\method therefore performs cross-request pruning over linear draft blocks and selects the pruning depth with a profiled runtime cost model, making the allocation adaptive to both batch confidence and deployment hardware.

\end{document}